\documentclass{article}
\PassOptionsToPackage{numbers, sort, compress}{natbib}
\usepackage[final, main]{neurips_2026}

\usepackage[english]{babel}
\usepackage{microtype}

\usepackage{graphicx}
\usepackage{amsmath}
\usepackage{amssymb}
\usepackage{booktabs}
\usepackage{hyperref}
\usepackage{url}
\usepackage{glossaries}
\usepackage{caption}
\usepackage{wrapfig}
\usepackage{xcolor}
\makeglossaries

\newacronym{llms}{LLMs}{Large Language Models}
\newacronym{llm}{LLM}{Large Language Model}
\newacronym{tda}{TDA}{Topological Data Analysis}
\newacronym{rag}{RAG}{Retrieval-Augmented Generation}
\newacronym{mteb}{MTEB}{Massive Text Embedding Benchmark}
\newacronym{bert}{BERT}{Bidirectional Encoder Representations from Transformers}
\newacronym{cka}{CKA}{Centered Kernel Alignment}
\newacronym{uts}{UTS}{Unified Topological Signatures}
\newacronym{pca}{PCA}{Principal Component Analysis}

\begin{document}
\title{From Topology to Retrieval: \\ Decoding Embedding Spaces with Unified Signatures}
\author{%
  Florian Rottach$^{*1,2}$ \And
  William Rudman$^{*3}$ \And
  Bastian Rieck$^{4}$ \And
  Harrisen Scells$^{1}$ \And
  Carsten Eickhoff$^{1}$ \\[0.5em]
  $^1$University of Tübingen \quad
  $^2$Boehringer Ingelheim GmbH, Biberach (Riss), Germany \\
  $^3$The University of Texas at Austin \quad
  $^4$Fribourg University \\
  \texttt{florian.rottach@boehringer-ingelheim.com}
}

\maketitle

\begin{abstract}
Studying the shape of text embedding spaces enhances our understanding 
of model behavior and how geometry affects downstream task performance. 
While a range of topological and geometric metrics have been explored, 
they are typically analyzed in isolation, and the connections between 
them remain largely unknown. In this paper, we present \textbf{Unified Topological Signatures (UTS)} for characterizing the 
structure of embedding spaces. We evaluate our method across a wide 
range of text embedding models and retrieval datasets, finding that it outperforms 
individual metrics in downstream performance prediction. Through several 
experiments, we demonstrate that the signatures characterize
embedding spaces more accurately, improve the prediction of model-specific properties and biases, and reveal novel insights into representational similarity 
that challenge the global claims of the Platonic Representation Hypothesis. \end{abstract}
\begin{figure}[h]
    \includegraphics[width=\textwidth]{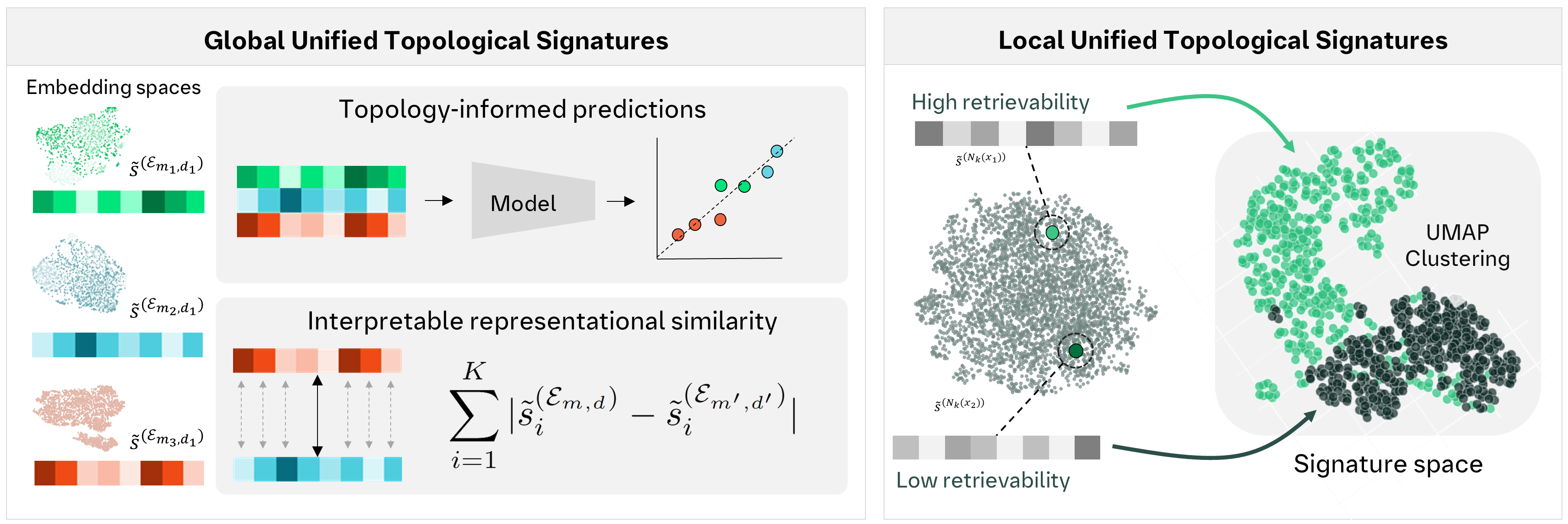}
    \caption{\textit{Unified Topological Signatures} of embedding spaces. Left: We construct signature vectors for embedding spaces by computing various topological descriptors. We use the vectors for downstream prediction tasks as well as for measuring representational similarity. Right: We compute local signatures based on the neighborhood of individual embeddings and use them to detect retrievability bias in document corpora.}
    \label{icml-historical}
\end{figure}

\section{Introduction}

\glsresetall
Text embeddings from transformer models are central for many tasks ranging from semantic search~\cite{zhao2024dense} to text classification~\cite{da2023text} and clustering~\cite{petukhova2025text}. They often play a crucial role in \gls*{rag}~\cite{lewis2020retrieval}, which combines parametric knowledge of a \gls*{llm} with external knowledge from a search engine, allowing \gls*{llm}s to respond with information not present in their training data. Dense retrieval models (i.e., architectures trained so that query embeddings are highly similar to embeddings of relevant documents) are a natural choice for \gls*{rag} and other information retrieval tasks due to their speed and effectiveness. Selecting the \emph{best} model for such a task remains an open challenge. While benchmarks such as \gls*{mteb}~\cite{muennighoff2023mtebmassivetextembedding} and theoretical frameworks~\cite{darrin2024embedding, caspari2024beyond} provide valuable tools for model comparison and selection, they do not predict or explain effectiveness on novel tasks. To gain deeper insight into representations, prior work has analyzed the geometry of word embeddings \cite{devlin2019bert} and internal representations of \gls*{llm}s \cite{gardinazzi2025persistent}. Notably, studies have shown that effective \gls*{llm} embeddings often exhibit \emph{anisotropy} \cite{razzhigaev2023shape}, with representations concentrating along a few principal axes. Additionally, evidence from the Platonic Representation Hypothesis~\cite{huh2024platonic} points to geometry as a key driver of model behavior, suggesting that representations from different models are converging. While these studies offer valuable insights, representational attributes are often analyzed in isolation and text embedding models remain underexplored, which limits our understanding of how geometry affects downstream task performance. 

In this paper, we address this gap by investigating the high-dimensional geometry of text embedding spaces to better understand what drives model effectiveness. Our main \textbf{contributions} are as follows:
\begin{enumerate}\setlength{\itemsep}{0ex}

\item We systematically study the combination of multiple geometric descriptors through \emph{Unified Topological Signatures} (UTS), showing that multidimensional signatures outperform individual metrics in predicting retrieval effectiveness and retrievability bias.

\item We demonstrate that embedding geometry is fundamentally multidimensional, with UTS providing distinctive geometric fingerprints of models and datasets.

\item We evaluate the global alignment claims of the Platonic Representation Hypothesis using UTS-based similarity. Rather than convergence by scale, we observe clustering by model family, highlighting the influence of architecture and training data on global embedding-space geometry.
\end{enumerate}
\section{Related Work}

\paragraph{Geometry of Embeddings}
 
Early studies have explored the internal representations of BERT~\cite{devlin2019bert}, focusing on how their geometric properties influence word senses~\cite{reif2019visualizing}. Other research has investigated isotropy~\cite{ethayarajh2019contextual}, isometry~\cite{Draganov_2024}, clustering patterns, and intrinsic dimensions~\cite{cai2021isotropy, torregrossa2020correlation} of contextualized word embeddings. In evaluation studies, geometry has been identified as one of the key properties driving embedding quality~\cite{wang2019evaluating}. Recently, attention has shifted towards analyzing the internal representations of \gls*{llm}s, which have been shown to be effective at many tasks without any fine-tuning~\cite{tao2025llmseffectiveembeddingmodels, gruver2024largelanguagemodelszeroshot}. Several works have studied the training dynamics of isotropy~\cite{razzhigaev2023shape, rudman2024stableanisotropicregularization} and intrinsic dimension~\cite{janapati2024comparative, cheng2405emergence, ruppik2025less}, as well as the topological differences across layers~\cite{gardinazzi2025persistent, liu2025spectral}. Recently,~\citet{lee2024geometric} found a connection between linguistic compositionality and the complexity of data manifolds. Theoretical work by ~\citet{weller2025theoretical} suggests that dimensionality is a key driver for the effectiveness of embedding models. While a large body of research is focused on sentence or generally text representations, others have analyzed intrinsic dimension and other properties specifically for token embeddings~\cite{kataiwa2025measuring, viswanathan2025geometry}. Notably,~\citet{lee2025shared} found that token embeddings across different models share a common geometric structure.  Unlike prior works examining individual geometric metrics, we present a \emph{unified} representation that incorporates the full topological profile into a signature vector, offering a more stable and rigorous summary.

\paragraph{Embedding-based Retrieval}

Dense retrieval has emerged as a powerful alternative to traditional lexical-based retrieval methods like BM25~\cite{robertson2009probabilistic}, which are based on sparse vector representations~\cite{zhao2024dense}. In recent years, multiple attempts have been made to understand the geometry of embedding spaces in the context of retrieval, such as investigating the influence of local intrinsic dimension on nearest neighbor search~\cite{aumuller2019role} and hierarchical navigable small worlds (HNSW) algorithms~\cite{elliott2024impacts}. In addition, theoretical work has shown that the embedding space dimensionality is a bounding factor for retrieval effectiveness~\cite{weller2025theoretical}. However, a more complete topological perspective has not been considered for retrieval performance prediction. In addition, we study the retrievability bias~\cite{wilkie2014best} in dense retrieval systems, which describes the phenomenon that some documents are easier to retrieve than others. Studies have found that larger bias corresponds to poorer retrieval performance \cite{wilkie2014best}, hence identifying such documents can improve retrieval systems through techniques like document augmentation~\cite{su2025parametricretrievalaugmentedgeneration} and other optimization strategies~\cite{setty2024improvingretrievalragbased}. Currently, no works have attempted to link retrievability bias to geometric properties of the embedding space. We show that we can predict whether or not a document is retrievable, indicating the importance of the geometric analysis of representations of dense retrieval models. 

\paragraph{Representational Similarity and Alignment}

From early work on neural pattern similarity to recent advances in cross-modal alignment in transformers, representational similarity has been central to machine learning~\cite{sucholutsky2023aligned}. There exist 
a variety of methods to compare representations, each assessing similarity with distinct aspects of granularity, symmetry, or sensitivity. In recent years, \gls*{cka}~\cite{kornblith2019similarity}, has been widely adopted for computing similarity, particularly due to its robustness under orthogonal transformations and isotropic scaling. ~\citet{caspari2024beyond} use CKA to identify architecture as a key driver for embedding similarity, while ~\citet{ciernikobjective} find the objective function determining representational consistency for vision models. The similarity of representations is also at the core of the \emph{Platonic Representation Hypothesis}~\cite{huh2024platonic}, which argues that representations converge and foundation models become more aligned. The findings of this study indicate that the degree of alignment is shaped by the choice of similarity metric, underscoring its critical role in evaluating representational consistency. While CKA is a powerful statistical tool, it may miss topological differences in embedding spaces ~\cite{davari2022reliability}, particularly on a global scale. Recent work has started to address this gap ~\cite{islam2025manifoldapproximationleadsrobust, barannikov2022representationtopologydivergencemethod}, but these methods focus on isolated aspects and offer little explanation for why representations diverge.

To address fundamental limitations of individual metrics, we investigate the potential of multidimensional signatures that summarize topology, intrinsic dimensionality, clustering behavior, and other measurable aspects of embedding spaces. In contrast to current similarity methods, we can pinpoint in what aspects embedding spaces differ. Furthermore, by comparing the global topology, we are not restricted to paired embedding spaces, which is a requirement for most representational similarity methods. Finally, we examine whether the Platonic Representation Hypothesis holds under our UTS-based similarity measure, providing evidence that global alignment is shaped by other factors than by model scale.

\section{Experimental Setup and Methodology}
\subsection{Unified Topological Signatures}

\paragraph{Definition} 
To provide a multidimensional characterization of embedding spaces, we aggregate topological and geometric descriptors into a vector representation. In the following, we define global and local variants, outline normalization, and introduce an interpretable similarity measure.

Let $T_i$ be a global descriptor of a point cloud in an $n$-dimensional real-valued space $X \subset \mathbb{R}^n$, where $i \in \{1, \dotsc, K\}$ and $K$ is the number of descriptors. Each descriptor is a real-valued function that captures a specific topological or geometric property of the point cloud, i.e.,
\begin{equation}
T_i\colon X \rightarrow s_i^X \in \mathbb{R}
\end{equation}
\glsreset{uts}%
Let $ \mathcal{E}_{m,d} \subset \mathbb{R}^n $ be the $n$-dimensional embedding space, generated by model \( m \in \mathcal{M} \) on dataset \( d \in \mathcal{D} \). We define the global \gls*{uts} of $\mathcal{E}_{m,d}$ as a vector:
\begin{equation}
    \boldsymbol{s}^{\mathcal{E}_{m,d}} = [T_1(\mathcal{E}_{m,d}), T_2(\mathcal{E}_{m,d}), \dots, T_K(\mathcal{E}_{m,d})]^\top
    \label{global_signature}
\end{equation}
where each component \( \boldsymbol{s}^{\mathcal{E}_{m,d}}_i \) corresponds to the value of distinct topological descriptor $T_i \in T$, with $i \in \{1, ..., K\}$. Table~\ref*{descriptors} summarizes the selected descriptors, which cover a wide range of properties. Details and mathematical definitions for each descriptor are provided in Appendix \ref*{a_desc}.

Each descriptor \( T_i \) is computed over a varying sample of \( n_i \) data points, drawn from \( \mathcal{E}_{m,d} \), with \( n_i \) chosen empirically to balance descriptor robustness and computational feasibility. This ensures that each descriptor captures meaningful topological features while being computationally tractable.

\paragraph{Local Signatures}
We define a local variant of the signatures of a point \( \boldsymbol{x} \in \mathcal{E}_{m,d} \) as:
\begin{equation}
\label{local_sign}
\boldsymbol{s}^{\mathcal{N}_k(x)} = [T_1(\mathcal{N}_k(x)), \dots, T_K(\mathcal{N}_k(x))]^\top
\end{equation}
where each component \( \boldsymbol{s}^{\mathcal{N}_k(x)}_i \) corresponds to the value of \( T_i \) computed over the 
$k$-nearest neighbors of \( \boldsymbol{x} \) denoted by \( \mathcal{N}_k(\boldsymbol{x}) \subset \mathcal{E}_{m,d} \).

\paragraph{Normalization}
To enable comparability across models and datasets, we normalize each signature vector computed for a point cloud $X$ by dividing its components by the maximum absolute observed value across all embedding spaces. Other normalization variants are evaluated in Appendix \ref*{a:ablations}.
\begin{equation}
\tilde{s}_i^{X} = \frac{s_i^{X}}{\max_{X'} |s_i^{(X')}|}
\quad \text{for } i = 1, \dots, K
\end{equation}
yielding the normalized signature vector:
\begin{equation}
\tilde{\boldsymbol{s}}^{X} = [\tilde{s}_1^{X}, \tilde{s}_2^{X}, \dots, \tilde{s}_K^{X}]^\top
\end{equation}
Instead of normalizing across all signatures, this normalization can be applied to a subgroup of signatures, such as vectors computed on a specific dataset.

\paragraph{Similarity}
Given two normalized signature vectors \( \tilde{\boldsymbol{s}}^{\mathcal{E}_{m,d}} \) and \( \tilde{\boldsymbol{s}}^{\mathcal{E}_{m',d'}}\), the similarity between the corresponding embedding spaces regarding property $i$ is quantified using the Manhattan distance:
\begin{equation}
\boldsymbol{\delta_i}^{(\mathcal{E}_{m,d}),(\mathcal{E}_{m',d'})} = \left| \tilde{\boldsymbol{s_i}}^{(\mathcal{E}_{m,d})} - \tilde{\boldsymbol{s_i}}^{(\mathcal{E}_{m',d'})} \right|
\end{equation}

We define a global similarity measure as the sum of these differences:
\begin{equation}
    d(\tilde{\boldsymbol{s}}^{(\mathcal{E}_{m,d})}, \tilde{\boldsymbol{s}}^{(\mathcal{E}_{m',d'})}) = \sum_{i=1}^{K} | \tilde{s}_i^{(\mathcal{E}_{m,d})} - \tilde{s}_i^{(\mathcal{E}_{m',d'})} |
    \label{similarity}
\end{equation}

We choose the Manhattan distance because it quantifies absolute differences without applying transformations and allows for a decomposition of element‑wise contributions to embedding‑space divergence, which is not possible with metrics such as Cosine similarity.

\subsection{Datasets}
Due to the increasing adoption of dense retrieval and its widespread applications in \gls*{rag} we focus our analysis on the retrieval datasets shown in Table~\ref*{datasets}, which are included in the MTEB evaluation framework. Formally, let $q$ denote a natural language query and $d_i$ denote a document from a large collection $D = \{d_i\}_{i=1}^m$ consisting of $m$ documents. Given a query, the goal of text retrieval is to return a ranked list of the $n$ most relevant documents $L = [d_1, d_2, \ldots, d_n]$ according to the relevance scores produced by a retrieval model. Dense retrieval models represent both queries and documents as dense vectors in a shared semantic space. The relevance score between a query and a document is then computed using a similarity function between their dense representations:
\[
    \mathrm{Rel}(q, d) = f_{\mathrm{sim}}(\phi_q(q), \phi_d(d))
\] 
where $\phi_q(\cdot)$ and $\phi_d(\cdot)$ are neural encoders mapping queries and documents into $n$-dimensional vectors, and $f_{\mathrm{sim}}$ denotes the similarity function. We conduct our experiments on 11 datasets, spanning multiple domains and corpus sizes, as detailed in Table~\ref*{datasets}.

\begin{figure}[t]
    \centering
    \begin{minipage}[b]{0.48\columnwidth} 
        \centering
        \tiny 
        \setlength{\tabcolsep}{2pt} 
        \renewcommand\arraystretch{0.8}
        
        \begin{tabular*}{\linewidth}{@{\extracolsep{\fill}}llrl@{}}
        \toprule
        \textsc{Measure} & \textsc{Category} & \textsc{\#Samp.} & \textsc{Src.} \\
        \midrule
        PH Dimension         & Homology  & 5K   & \cite{adams2020fractal} \\
        PH Statistics         & Homology  & 5K   & \cite{edelsbrunner2008persistent} \\
        Persistence Entropy    & Homology  & 20K  & \cite{chintakunta2015entropy} \\
        Euler Characteristic      & Homology  & 5K   & \cite{basu1996bounding} \\
        TwoNN Dimension       & Intrinsic Dimension  & 50K  & \cite{facco2017estimating} \\
        PCA Dim.         & Intrinsic Dimension  & 50K  & \cite{fukunaga1971algorithm} \\
        Effective Rank        & Intrinsic Dimension  & 100K & \cite{roy2007effective} \\
        Magnitude Dimension        & Diversity  & 5K   & \cite{leinster2013asymptotic} \\
        Magnitude Area        & Diversity  & 5K   & \cite{limbeck2025metricspacemagnitudeevaluating} \\
        Spread           & Diversity  & 10K  & \cite{willerton2015spread} \\
        Vendi Score      & Diversity  & 20K  & \cite{friedman2022vendi} \\
        Pairwise Similarity    & Density & 50K  & N/A \\
        Uniformity       & Uniformity & 20K  & \cite{wang2020understanding} \\
        IsoScore         & Isotropy & 500K & \cite{rudman2021isoscore} \\
        Silhouette       & Clustering & 20K  & \cite{shahapure2020cluster} \\
        \bottomrule
        \end{tabular*}
        \captionsetup{type=table}
        \caption{Descriptors for signature vectors.}
        \label{descriptors}
        
        \vspace{10pt} 
        
        \begin{tabular*}{\linewidth}{@{\extracolsep{\fill}}llrrr@{}}
        \toprule
        \textsc{Dataset} & \textsc{Domain} & \textsc{\#Q} & \textsc{\#Docs} & \textsc{Src.} \\
        \midrule
        MS MARCO         & Web    & 27    & 8.8M  & \cite{nguyen2016ms} \\
        ArguAna          & Argument   & 1.4K  & 8.7K  & \cite{wachsmuth2018retrieval} \\
        Touche2020       & Argument   & 249   & 528K  & \cite{thakur2021beir} \\
        NFCorpus         & Medical   & 323   & 3.6K  & \cite{boteva2016} \\
        TREC-COVID       & Medical   & 50    & 171K  & \cite{voorhees2020treccovidconstructingpandemicinformation} \\
        SciFact          & Science   & 300   & 5.2K  & \cite{wadden2020factfictionverifyingscientific} \\
        SCIDOCS          & Science   & 1K    & 25.6K & \cite{cohan2020specterdocumentlevelrepresentationlearning} \\
        ClimateFEVER     & Science   & 154   & 5.4M  & \cite{diggelmann2021climatefever} \\
        CQADupstack      & Forum  & 1.6K  & 40K   & \cite{hoogeveen2015cqadupstack} \\
        QuoraRetrieval   & Forum  & 10K   & 523K  & \cite{thakur2021beir} \\
        FiQA2018         & Finance   & 648   & 57K   & \cite{maia2018financial} \\
        \bottomrule
        \end{tabular*}
        \captionsetup{type=table}
        \caption{Summary of retrieval datasets.}
        \label{datasets}
    \end{minipage}
    \hspace{1.5em}
    \begin{minipage}[b]{0.4\columnwidth}
        \centering
        \scriptsize
        \renewcommand\arraystretch{0.7}
        \begin{tabular}{lrrl}
        \toprule
        \textsc{Model} & \textsc{Dim.} & \textsc{Size} & \textsc{Src.} \\
        \midrule
        gte-Qwen2-1.5B       & 1536 & 1.5B  & \citenum{li2023towards} \\
        gte-Qwen2-7B         & 3584 & 7B    & \citenum{li2023towards} \\
        stella-en-1.5B-v5    & 1024 & 1.5B  & \citenum{zhang2025jasperstelladistillationsota} \\
        Qwen3-Emb-0.6B       & 1024 & 0.6B  & \citenum{qwen3embedding} \\
        Qwen3-Emb-4B         & 2560 & 4B    & \citenum{qwen3embedding} \\
        Qwen3-Emb-8B         & 4096 & 8B    & \citenum{qwen3embedding} \\
        Linq-Emb-Mistral   & 4096 & 7.3B  & \citenum{LinqAIResearch2024} \\
        SFR-Emb-Mistral      & 4096 & 7.3B  & \citenum{SFRAIResearch2024} \\
        e5-mistral-7b        & 4096 & 7.3B  & \citenum{wang2022text} \\
        bge-small-en-v1.5    & 384  & 0.02B & \citenum{bge_embedding} \\
        bge-base-en-v1.5     & 768  & 0.1B  & \citenum{bge_embedding} \\
        bge-large-en-v1.5    & 1024 & 0.3B  & \citenum{bge_embedding} \\
        me5-large-instruct   & 1024 & 0.56B & \citenum{wang2024multilingual} \\
        arctic-embed-m-v1.5  & 768  & 0.1B  & \citenum{merrick2024arctic} \\
        ember-v1.0           & 1024 & 0.3B  & \citenum{nur2024emberv1} \\
        mxbai-xsmall-v1      & 384  & 0.1B  & \citenum{xsmall2024mxbai} \\
        mxbai-large-v1       & 512  & 0.3B  & \citenum{emb2024mxbai} \\
        all-MiniLM-L6-v2     & 384  & 0.02B & \citenum{wang2020minilmdeepselfattentiondistillation} \\
        gtr-t5-base          & 768  & 0.1B  & \citenum{ni2021largedualencodersgeneralizable} \\
        gtr-t5-large         & 768  & 0.3B  & \citenum{ni2021largedualencodersgeneralizable} \\
        gtr-t5-xxl           & 768  & 4.8B  & \citenum{ni2021largedualencodersgeneralizable} \\
        gte-small            & 384  & 0.3B  & \citenum{li2023towards} \\
        gte-base             & 768  & 0.1B  & \citenum{li2023towards} \\
        gte-large            & 1024 & 0.3B  & \citenum{li2023towards} \\
        msmarco-roberta      & 768  & 0.1B  & \citenum{xiong2020approximate} \\
        msmarco-distilbert   & 768  & 0.06B & \citenum{Hofstaetter2021_tasb_dense_retrieval} \\
        gemini-emb-001       & 3072 & N/A   & \citenum{lee2025geminiembeddinggeneralizableembeddings} \\
        openai-emb-3-small   & 1536 & N/A   & \citenum{openai2023embedding} \\
        openai-emb-3-large   & 3072 & N/A   & \citenum{openai2023embedding} \\
        \bottomrule
        \end{tabular}
        \captionsetup{type=table}
        \caption{Embedding models.}
        \label{models}
    \end{minipage}
\end{figure}

\subsection{Retrievability bias}
A document's retrievability \cite{wilkie2014best} is defined as the ease with which it can be retrieved by a given retrieval system across a set of general queries. More formally, the retrievability score $r(d)$ of a document $d$ is defined as:
\begin{equation*}
    r(d) \propto \sum_{q \in Q} f(k_{dq}, c)
\end{equation*}
where $Q$ is a set of queries, $k_{dq}$ is the rank of document $d$ in the retrieval results for query $q$, $c$ is a cutoff rank (e.g., top-10), and $f(k_{dq}, c)$ is a function that assigns higher weights to documents appearing at higher ranks. A common choice for $f$ is
\begin{equation*}
f(k_{dq}, c)=
\begin{cases}
1 & \textrm{if $d$ is in the top $c$ documents given $q$},\\
0 & \textrm{otherwise}.
\end{cases}
\end{equation*}

The retrievability of a document is primarily determined by its similarity to a query. For most datasets, documents with a high query alignment (i.e., those that closely match the terms or intent of typical queries)
will be more retrievable than documents that lie outside the typical query space. In dense retrieval systems documents and queries are embedded into a shared space in which similarity is influenced by the choice of embedding model. MTEB reveals that retrievability varies across embedding models despite a constant set of queries, underscoring the influence of the embedding space structure. Therefore, we expect significant differences in the topological patterns of different documents, driven by linguistic and semantic properties.

\subsection{Embedding models}
We use 27 text embedding models from different families, including Qwen \cite{qwen3embedding}, E5 \cite{wang2022text}, T5 \cite{ni2021largedualencodersgeneralizable}, BGE \cite{bge_embedding}, Gemini \cite{lee2025geminiembeddinggeneralizableembeddings} and others as listed in Table \ref*{models}. The models vary in terms of size, embedding dimension, architecture, and training data. We use the models' default text preprocessing and tokenization methods and normalize all embeddings to unit length. For Gemini, we specifically use the embeddings optimized for retrieval tasks. Across every model-dataset pair, we generate document and query embeddings using the \gls*{mteb} framework. Open-source models are downloaded and executed via HuggingFace \cite{wolf2020huggingfacestransformersstateoftheartnatural}, while embeddings for the OpenAI and Google models are produced through their respective APIs.

\subsection{Computation of signatures}
For each embedding space generated by model \( m \) on dataset \( d \), we compute a topological signature as defined by Equation \ref*{global_signature}. The descriptors cover a wide range of properties commonly found in other works, including homology, intrinsic dimension, diversity, density, isotropy, and clustering. For descriptors utilizing pairwise-distances, we compute both the Euclidean and Cosine distance functions. As some methods are intractable to calculate for millions of embeddings, we perform a preliminary analysis to assess the robustness of each descriptor with respect to sample size, as discussed in Appendix \ref*{a_scale}. We then sample $n_i$ datapoints for each descriptor \( T_i \) and compute the corresponding value. To produce a larger dataset with more variability, we repeat this procedure for 3 different seeds, such that we have 891 embedding space signatures for 27 models and 11 datasets. We then normalize each signature as described above. All computations were performed using parallel processing on 8 GPU compute nodes.

\subsection{Experiments}
\paragraph{Topology-informed downstream prediction}
We fit a Random Forest model from scikit-learn \cite{kramer2016scikit} to predict recall, mean average precision (MAP), and normalized discounted cumulative gain (NDCG) for all datasets using cutoff values of $k=\{5, 20, 100\}$ with the UTS vectors as input features. The retrieval measures have been computed using \gls*{mteb} \cite{muennighoff2023mtebmassivetextembedding}, and z-normalization is applied to make the metric scales comparable between datasets. Given the limited training data, we use 3-fold group cross-validation to predict on unseen datasets and evaluate the correlation using Spearman's $\rho$. We ensure that all data preprocessing steps are based on the training data distribution to avoid data leakage. To prevent overfitting, we limit the maximum depth of the decision trees to 5. We assess the prediction quality, by comparing UTS against the following single metric descriptors, frequently used in other works: TwoNN intrinsic dimension, effective rank, mean pairwise similarity and IsoScore.

To explore the effects of embedding space structure on retrievability bias in more detail, we compute the retrievability with $c=100$ for all documents in a corpus using all available queries from the respective dataset. We then compute local signatures (see Equation \ref*{local_sign} with $k=100$) for the 100 most and least retrieved documents \( x \) in the embedding space. This results in 59400 signatures, which we use as features to fit a binary Random Forest classifier to predict whether a document is highly retrievable. We follow the same cross-validation strategy as for the global signatures and include several baselines: the local density (mean pairwise similarity), the local intrinsic dimension, which has been found to influence retrieval \cite{aumuller2019role} as well as the linguistic feature word count, which has been linked to retrievability in previous studies \cite{wilkie2013relating}. In addition, we run a baseline experiment predicting retrievability directly from the raw embeddings. Finally, we analyze the distribution of the retrievability scores to identify the overall retrievability bias. To measure the \textit{inequality} of the distribution, we compute the Gini coefficient \cite{wilkie2014efficiently} and study the differences between models.

\paragraph{Characterizing embedding spaces}
We compare UTS against individual descriptors in their ability to separate embedding spaces. For this, we cross-validate a Random Forest classifier to predict the embedding model and model family corresponding to an unseen embedding space. This allows us to measure how discriminative the topological signatures are. In addition, we investigate whether model-specific information is encoded in the geometry of the embedding space by predicting the architecture and activation function used in text embedding models. This approach is particularly interesting for proprietary models, where such information is not publicly available. We include random predictions and the most frequent class as trivial baselines. We further evaluate individual topology descriptors, namely the TwoNN intrinsic dimension, effective rank, average pairwise similarity, and IsoScore. To investigate which properties contribute most to topological variation, we perform a Principal Component Analysis (PCA) on the global UTS vectors and analyze correlations between the individual descriptors. We also include a PCA-transformed version of UTS as an additional baseline. Since some properties are affected by class imbalance, we use balanced accuracy as the evaluation metric.

\paragraph{Topological clustering}
We test the Platonic Representation Hypothesis by analyzing whether embedding space geometry converges with increasing scale. For this, we compute global \gls*{uts} distances between models according to Equation \ref*{similarity}. We average the distances between models across all datasets, yielding a 27x27 distance matrix. To confirm whether larger models have more similar topologies, we group models into the categories small, medium and large using the parameter count thresholds of $<0.5$B, $0.5-3$B and $>3$B parameters, respectively. To test alternative drivers of representational similarity, we also group the distances by model family, investigating if architectural properties and training data explain the differences. 
The source code to generate the embedding spaces and reproduce our experiments is available at \url{https://github.com/Boehringer-Ingelheim/unified-topological-signatures}.
\section{Results}

\paragraph{No single metric tells the whole story.}

\begin{figure}[t] \centering \setlength{\abovecaptionskip}{2pt} \setlength{\belowcaptionskip}{-10pt} \includegraphics[width=0.55\columnwidth]{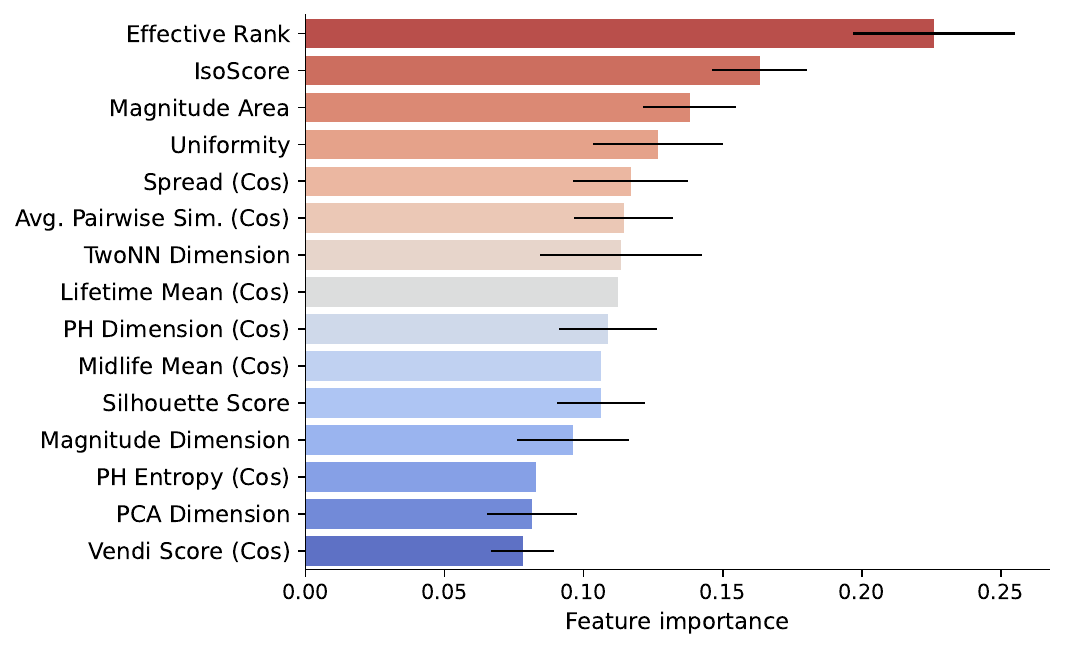} \caption{Feature importance across CV-folds.} \label{imp} \end{figure}

As summarized by Table \ref*{performance-pred}, we find that UTS significantly outperform individual metrics in retrieval performance prediction. The best performance is achieved for Recall@20 with a mean Spearman correlation of $0.70 \pm 0.11$ across the cross-validation folds. The best single metric is effective rank with a correlation of $0.47 \pm 0.08$. These results highlight the limitations of relying on any single geometric descriptor. Our multi-attribute perspective, by contrast, captures more complex structural patterns, translating into substantially stronger predictive power. To better understand these patterns, we analyze the feature importance across all folds, summarized in Figure \ref*{imp} and find that the effective rank is the most important feature. This result is in line with recent theoretical work \cite{weller2025theoretical} suggesting that retrieval performance is limited by the dimensionality of the embeddings. Notably, we find that other descriptors also play a significant role, highlighting that the predictive signal is distributed across the full signature. We discuss more results and experiments for retrieval performance prediction in Appendix \ref*{a_retrieval}.

\begin{table}[t]
\caption{Spearman's $\rho$ for retrieval performance prediction.}
\label{performance-pred}
\centering
\small
\setlength{\tabcolsep}{2.5pt}
\renewcommand{\arraystretch}{1}
\begin{tabular}{lcccccccccc}
\toprule
 & \multicolumn{3}{c}{Recall} & \multicolumn{3}{c}{MAP} & \multicolumn{3}{c}{NDCG} \\
\cmidrule(lr){2-4} \cmidrule(lr){5-7} \cmidrule(lr){8-10}
Method & @5 & @20 & @100 & @5 & @20 & @100 & @5 & @20 & @100 \\
\midrule
TwoNN           & $0.07_{\pm.09}$ & $0.08_{\pm.11}$ & $0.04_{\pm.09}$ & $0.07_{\pm.11}$ & $0.06_{\pm.11}$ & $0.04_{\pm.09}$ & $0.06_{\pm.11}$ & $0.07_{\pm.10}$ & $0.05_{\pm.09}$ \\
Eff.\ rank      & $0.44_{\pm.11}$ & $0.47_{\pm.08}$ & $0.42_{\pm.09}$ & $0.44_{\pm.09}$ & $0.43_{\pm.08}$ & $0.43_{\pm.07}$ & $0.45_{\pm.09}$ & $0.45_{\pm.08}$ & $0.45_{\pm.07}$ \\
Mean sim.       & $0.17_{\pm.13}$ & $0.18_{\pm.15}$ & $0.20_{\pm.11}$ & $0.17_{\pm.11}$ & $0.18_{\pm.12}$ & $0.19_{\pm.13}$ & $0.20_{\pm.13}$ & $0.19_{\pm.14}$ & $0.19_{\pm.14}$ \\
IsoScore        & $0.23_{\pm.15}$ & $0.34_{\pm.06}$ & $0.30_{\pm.07}$ & $0.24_{\pm.10}$ & $0.24_{\pm.11}$ & $0.25_{\pm.10}$ & $0.24_{\pm.10}$ & $0.27_{\pm.09}$ & $0.26_{\pm.10}$ \\
\midrule
UTS-PCA        & $0.49_{\pm.03}$ & $0.56_{\pm.05}$ & $0.55_{\pm.10}$ & $0.49_{\pm.01}$ & $0.49_{\pm.01}$ & $0.50_{\pm.02}$ & $0.48_{\pm.03}$ & $0.52_{\pm.01}$ & $0.52_{\pm.02}$ \\
\textbf{UTS}          & $\mathbf{0.64}_{\pm.16}$ & $\mathbf{0.70}_{\pm.11}$ & $\mathbf{0.67}_{\pm.12}$ & $\mathbf{0.63}_{\pm.17}$ & $\mathbf{0.66}_{\pm.14}$ & $\mathbf{0.68}_{\pm.15}$ & $\mathbf{0.66}_{\pm.14}$ & $\mathbf{0.69}_{\pm.13}$ & $\mathbf{0.68}_{\pm.14}$ \\
\bottomrule
\end{tabular}
\end{table}

\paragraph{Local topological signatures predict document retrievability.}

\begin{figure}[h]
    \centering
    \includegraphics[width=0.4\columnwidth]{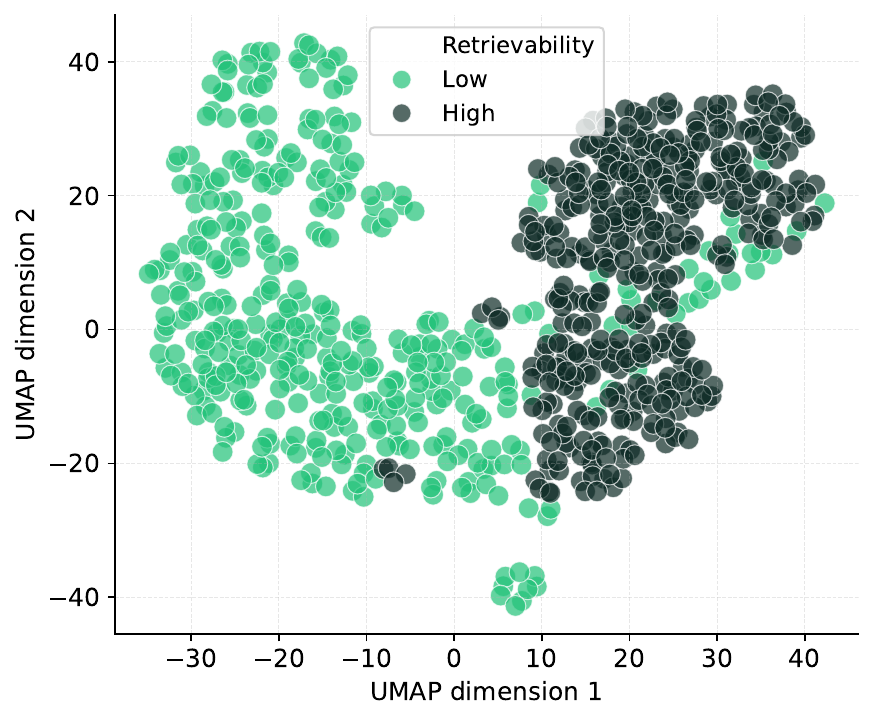}
    \hfill
    \includegraphics[width=0.49\columnwidth]{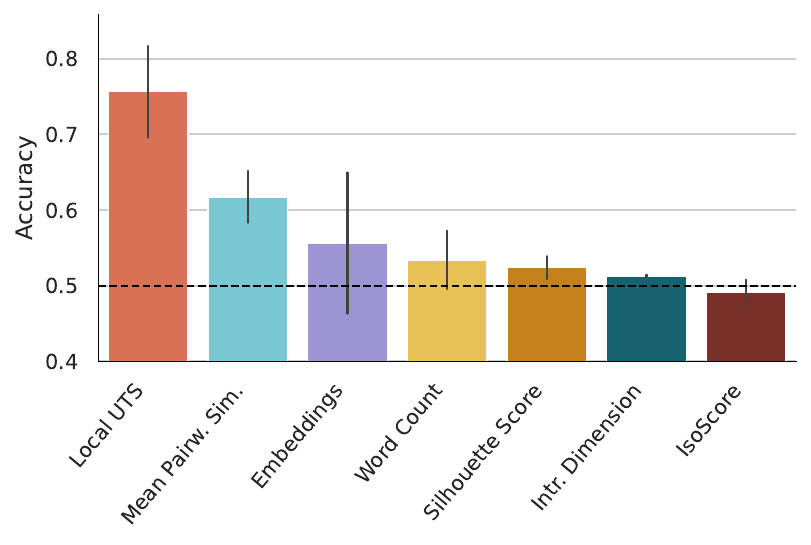}
    \\[4pt]
    \begin{minipage}[t]{0.49\columnwidth}
        \caption{UMAP visualization of local \gls*{uts} space for 100 highly retrievable and non-retrievable documents for QuoraRetrieval.}
        \label{umap-local}
    \end{minipage}
    \hfill
    \begin{minipage}[t]{0.49\columnwidth}
        \caption{Cross-validation accuracy for predicting document retrievability. The dashed line indicates a random baseline.}
        \label{predlocal}
    \end{minipage}
\end{figure}

Our analysis reveals that local \gls*{uts} serve as strong predictors of document retrievability, which can be a useful approach for identifying and mitigating bias in retrieval systems. We find that highly retrievable and non-retrievable documents are well separated in the local signature space, as visualized in Figure \ref*{umap-local} using UMAP \cite{mcinnes2018umap}. Figure \ref*{predlocal} summarizes the results of using signature vectors compared to baseline models that only use a single feature. We find that the signature-based model achieves an average accuracy of 0.76 across all dataset cross-validation folds, significantly outperforming the baselines. This demonstrates that retrievability detection requires a holistic analysis of the local topology, beyond simple one-dimensional measures. In Appendix \ref*{a_retrievability}, we provide additional analyses that highlight the structural differences in the embedding space between highly retrievable and non-retrievable documents. We also report model-specific accuracies, showing that OpenAI's text-embedding-3-large is most effective for the retrievability prediction task. Finally, we investigate the overall retrievability bias of embedding models using the Gini coefficient. This analysis reveals that decoder-based models exhibit greater bias (i.e., inequality in document exposure), across all datasets except ArguAna.

\paragraph{UTS uniquely characterize embedding space geometry.}

Table \ref*{tab:baseline-comparison} summarizes the results for predicting different model attributes using the \gls*{uts} as feature vectors. For unseen datasets, we can identify the specific model that generated an embedding space with a balanced accuracy of $0.66 \pm 0.02$ across all cross-validation folds, compared to a baseline accuracy of $0.04$, which corresponds to always predicting the most frequent model among the 27 classes. We perform the same analysis for predicting the model family, which is an easier task with an overall balanced accuracy of $0.68 \pm 0.07$ against a baseline accuracy of $0.17$. Our comparison with individual descriptors highlights that our multi-attribute approach significantly outperforms other methods for characterizing embedding spaces. Notably, effective rank is the best individual descriptor with a balanced accuracy of $0.39 \pm 0.02$ for predicting the model, which is around $40\%$ worse than our approach. Analyzing the feature importance shows that descriptors related to embedding dimension (e.g. effective rank) are most important for identifying a model. This strong discriminative power can be attributed to the fact that the embedding dimensions are tightly linked to model architecture, with some consistently exhibiting larger dimensions. To account for this, we run the experiments separately for each embedding dimension with sufficient data (Table \ref{pred-cv} in the Appendix) and find that the balanced accuracy increases to $0.89 \pm 0.10$ for 4096 dimensions, $0.68 \pm 0.23$ for 1024 dimensions, $0.73 \pm 0.12$ for 768 dimensions and $0.76 \pm 0.01$ for 384 dimensions. This shows that the \gls*{uts} can uniquely characterize embedding models, even when controlling for embedding dimension. We discuss the prediction of architecture and activation function in Appendix~\ref*{a_predictive}.

\begin{table}[t]
\centering
\small
\renewcommand{\arraystretch}{1.0}
\caption{Cross-validated balanced accuracy of UTS versus baselines.}
\label{tab:baseline-comparison}

\resizebox{0.8\columnwidth}{!}{
\begin{tabular}{lcccc}

Method & Model & Family & Architecture & Activation  \\
\midrule
Random      & 0.04 & 0.08 & 0.50 & 0.33 \\
Most frequent  & 0.04 & 0.17 & 0.72 & 0.59 \\
\midrule
Embeddings  & $0.26_{\pm.01}$ & $0.37_{\pm.02}$ & $0.78_{\pm.01}$ & $0.38_{\pm.01}$ \\
TwoNN dimension       & $0.02_{\pm.00}$ & $0.09_{\pm.01}$ & $0.02_{\pm.00}$ & $0.32_{\pm.01}$ \\
Effective rank   & $0.39_{\pm.02}$ & $0.47_{\pm.02}$ & $\mathbf{0.98}_{\pm.00}$ & $0.78_{\pm.02}$ \\
Mean similarity   & $0.23_{\pm.05}$ & $0.46_{\pm.04}$ & $0.63_{\pm.04}$ & $0.56_{\pm.02}$ \\
IsoScore    & $0.11_{\pm.03}$ & $0.16_{\pm.01}$ & $0.79_{\pm.03}$ & $0.53_{\pm.02}$ \\
\midrule
UTS-PCA     & $0.44_{\pm.03}$ & $0.50_{\pm.05}$ & $0.97_{\pm.01}$ & $0.83_{\pm.07}$ \\
\textbf{UTS} & $\mathbf{0.66}_{\pm.02}$ & $\mathbf{0.68}_{\pm.07}$ & $\mathbf{0.98}_{\pm.01}$ & $\mathbf{0.84}_{\pm.02}$ \\
\bottomrule
\end{tabular}
}
\end{table}

\begin{figure}[t]
    \centering
    \includegraphics[width=0.49\columnwidth]{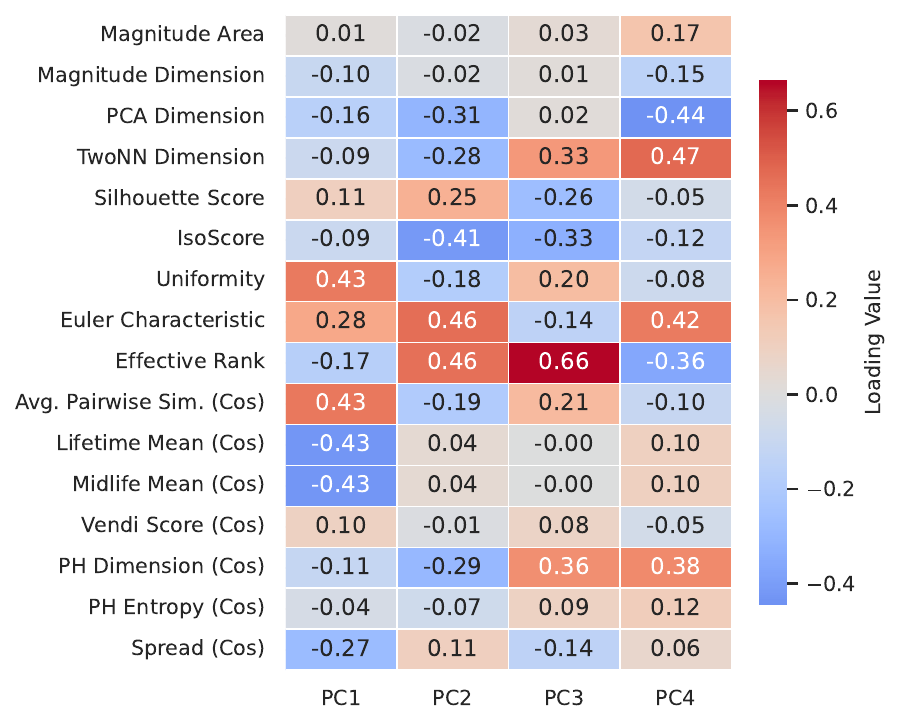}
    \hfill
    \includegraphics[width=0.49\columnwidth]{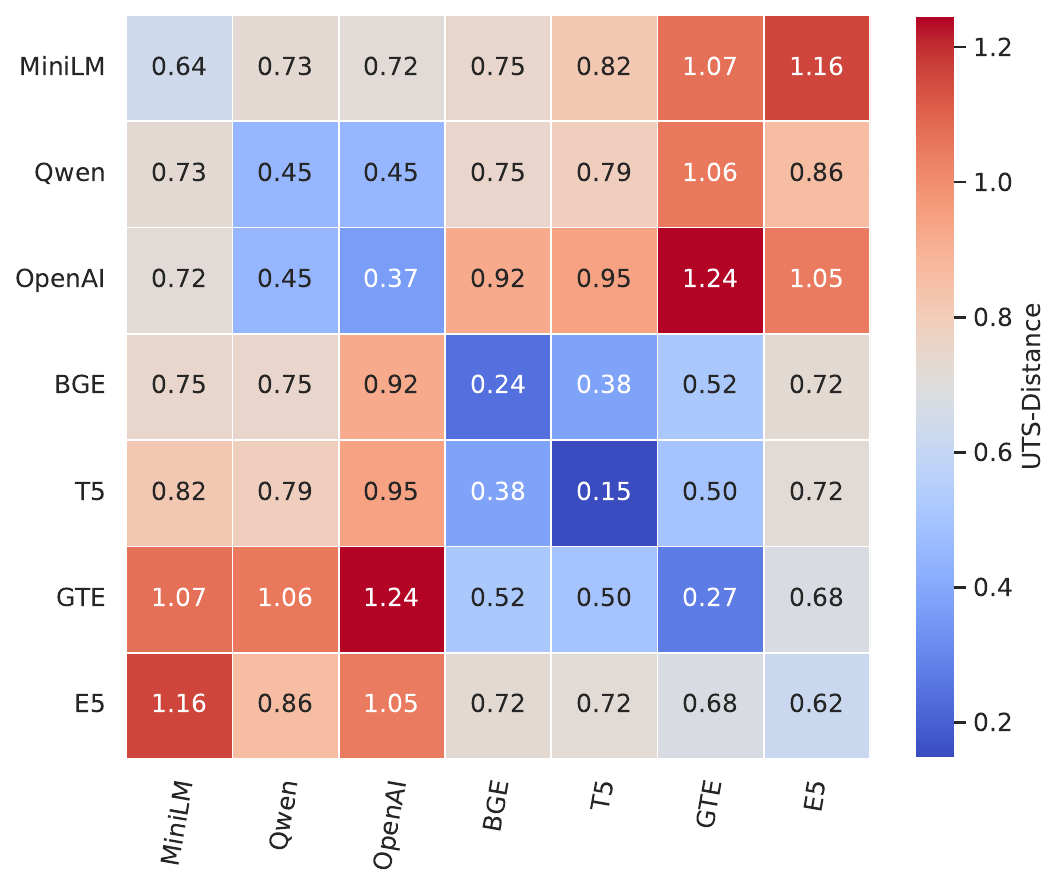}
    \\[4pt]
    \begin{minipage}[t]{0.49\columnwidth}
        \caption{PCA analysis of global UTS revealing that embedding space geometry can be summarized by only a few principal components.}
        \label{pca}
    \end{minipage}
    \hfill
    \begin{minipage}[t]{0.49\columnwidth}
        \caption{Model family dominates representational similarity, with lowest distances along the diagonal and clustering by architectures.}
        \label{fam_clust}
    \end{minipage}
\end{figure}

\paragraph{Global topology is summarized by a few principal components.}
The PCA reveals that $91\%$ of the variance is captured by the first four principal components, as shown in Figure \ref*{pca}. We inspect the loadings of each descriptor on the principal components and find that the first component mostly contains topological measures that are related to the pairwise distances between embeddings, including persistent homology features and kernel-based aggregations such as spread. The second component is dominated by measures related to dimensionality, isotropy, and clustering. These findings are supported by recent work establishing a connection between isotropy and clustering \cite{mickus2024isotropy}. The remaining components contain information on various intrinsic dimension measures, as well as the effective rank. In addition to this variability analysis, we present a correlation analysis between all descriptors and model-specific properties in Appendix \ref*{a_corr}. While we find that many descriptors are highly correlated, we observe a substantial performance gap between UTS-PCA and the full UTS representation, suggesting that a compressed set of principal components is not sufficient to capture all relevant topological information.

\paragraph{Global topologies cluster by model family.}
Linkage-based hierarchical clustering \cite{ackerman2016characterization} reveals that model family - not size - is the dominant factor shaping representational similarity between text embedding models. To summarize this, we aggregate inter-family distances by averaging off-diagonal entries in the upper triangular part of the matrix (see  Figure \ref*{fam_clust}). Intra-family distances are consistently lower, indicating that architecture and training data are more influential than scale. Furthermore, BERT-based models (BGE, T5, GTE) cluster tightly, as well as Qwen and OpenAI models, despite size variation within the family. These findings suggest that convergence in model representations is heavily shaped by architectural choices, challenging the assumption that scaling alone drives representational convergence. Our results are in line with recent work that shows that the Platonic Representation Hypothesis does not hold for global alignment \cite{groger2026revisiting}. In Appendix \ref*{a_similairty} we present additional experiments and compare our findings to similarity measured using \gls*{cka}.

\section{Discussion and Conclusion}
\paragraph{Discussion}
Our results emphasize the value of a multidimensional perspective on embedding space geometry, demonstrating clear advantages over individual metrics. We show that embedding space structure predicts downstream performance and bias on unseen tasks, enabling the assessment of embedding space quality without task-specific ground-truth labels. We highlight a practical application by accurately predicting document retrievability, enabling targeted bias mitigation through document augmentation. We further demonstrate that UTS characterize embedding spaces and recover internal model properties with high accuracy, significantly outperforming all baselines and offering a novel method to decode characteristics of proprietary architectures. Finally, our results challenge the Platonic Representation Hypothesis: rather than converging toward a global geometry as models scale, alignment is mainly driven by architectural choices and training data. For practitioners, incorporating anisotropic regularization and larger embedding dimensions improves structural quality, while our topological framework serves as a monitoring tool for embedding space quality.

\paragraph{Conclusion}
In this work, we introduce Unified Topological Signatures, enabling a multidimensional view of embedding space geometry. We compute a diverse set of topological and geometric descriptors for a wide range of embedding models and retrieval datasets. We demonstrated that these signatures can be used to predict model properties, retrieval performance, and retrievability bias. Furthermore, our analysis reveals that models from the same family and size category share the most similar topological properties, indicating that architecture and training data have a strong influence on the embedding space geometry. Our findings suggest that topological signatures provide a powerful tool for understanding and optimizing text embedding spaces for retrieval tasks and beyond. 

\paragraph{Limitations and Future work}
Our analysis is limited to retrieval tasks and English-language embeddings, leaving open how topological signatures behave in clustering, classification, or cross-lingual settings. Additionally, computing topological descriptors over large embedding spaces currently requires subsampling, which introduces approximation error and limits scalability. Addressing this would remove a practical bottleneck and enable analysis at larger scales. Beyond these limitations, further research is needed to understand how specific architectural choices and training data shape embedding space geometry. Finally, extending the framework to other data modalities would provide broader evidence for the relationship between topology and downstream task performance.

\section*{Acknowledgments} We thank I-Ting Ho, Jan Stanicki and Sebastian Schieferdecker for helpful discussions and feedback on this work. Computational resources were provided by the Boehringer Ingelheim High Performance Computing Cluster.

\clearpage

\bibliographystyle{plainnat}
\bibliography{icml26-signatures-lit}

@inproceedings{Draganov_2024,
   title={The Shape of Word Embeddings: Quantifying Non-Isometry with Topological Data Analysis},
   url={http://dx.doi.org/10.18653/v1/2024.findings-emnlp.705},
   DOI={10.18653/v1/2024.findings-emnlp.705},
   booktitle={Findings of the Association for Computational Linguistics: EMNLP 2024},
   publisher={Association for Computational Linguistics},
   author={Draganov, Ondřej and Skiena, Steven},
   year={2024},
   pages={12080–12099} }

@article{ethayarajh2019contextual,
  title={How contextual are contextualized word representations? Comparing the geometry of BERT, ELMo, and GPT-2 embeddings},
  author={Ethayarajh, Kawin},
  journal={arXiv preprint arXiv:1909.00512},
  year={2019}
}

@inproceedings{torregrossa2020correlation,
  title={On the correlation of word embedding evaluation metrics},
  author={Torregrossa, Fran{\c{c}}ois and Claveau, Vincent and Kooli, Nihel and Gravier, Guillaume and Allesiardo, Robin},
  booktitle={LREC 2020-12th Conference on Language Resources and Evaluation},
  pages={4789--4797},
  year={2020}
}

@misc{rudman2024stableanisotropicregularization,
      title={Stable Anisotropic Regularization}, 
      author={William Rudman and Carsten Eickhoff},
      year={2024},
      eprint={2305.19358},
      archivePrefix={arXiv},
      primaryClass={cs.CL},
      url={https://arxiv.org/abs/2305.19358}, 
}

@article{reif2019visualizing,
  title={Visualizing and measuring the geometry of BERT},
  author={Reif, Emily and Yuan, Ann and Wattenberg, Martin and Viegas, Fernanda B and Coenen, Andy and Pearce, Adam and Kim, Been},
  journal={Advances in neural information processing systems},
  volume={32},
  year={2019}
}

@inproceedings{cai2021isotropy,
  title={Isotropy in the contextual embedding space: Clusters and manifolds},
  author={Cai, Xingyu and Huang, Jiaji and Bian, Yuchen and Church, Kenneth},
  booktitle={International conference on learning representations},
  year={2021}
}

@article{razzhigaev2023shape,
  title={The shape of learning: Anisotropy and intrinsic dimensions in transformer-based models},
  author={Razzhigaev, Anton and Mikhalchuk, Matvey and Goncharova, Elizaveta and Oseledets, Ivan and Dimitrov, Denis and Kuznetsov, Andrey},
  journal={arXiv preprint arXiv:2311.05928},
  year={2023}
}

@inproceedings{
    gardinazzi2025persistent,
    title={Persistent Topological Features in Large Language Models},
    author={Yuri Gardinazzi and Karthik Viswanathan and Giada Panerai and Alessio ansuini and Alberto Cazzaniga and Matteo Biagetti},
    booktitle={Forty-second International Conference on Machine Learning},
    year={2025},
    url={https://openreview.net/forum?id=qAHnSkHvsm}
}

@article{janapati2024comparative,
  title={A Comparative Study of Learning Paradigms in Large Language Models via Intrinsic Dimension},
  author={Janapati, Saahith and Ji, Yangfeng},
  journal={arXiv preprint arXiv:2412.06245},
  year={2024}
}

@article{liu2025spectral,
  title={Spectral insights into data-oblivious critical layers in large language models},
  author={Liu, Xuyuan and Hsiung, Lei and Yang, Yaoqing and Yan, Yujun},
  journal={arXiv preprint arXiv:2506.00382},
  year={2025}
}

@article{cheng2405emergence,
  title={Emergence of a high-dimensional abstraction phase in language transformers, 2024},
  author={Cheng, Emily and Doimo, Diego and Kervadec, Corentin and Macocco, Iuri and Yu, Jade and Laio, Alessandro and Baroni, Marco},
  journal={URL https://arxiv. org/abs/2405.15471}
}

@article{kataiwa2025measuring,
  title={Measuring Intrinsic Dimension of Token Embeddings},
  author={Kataiwa, Takuya and Hakaze, Cho and Ohki, Tetsushi},
  journal={arXiv preprint arXiv:2503.02142},
  year={2025}
}

@article{viswanathan2025geometry,
  title={The geometry of tokens in internal representations of large language models},
  author={Viswanathan, Karthik and Gardinazzi, Yuri and Panerai, Giada and Cazzaniga, Alberto and Biagetti, Matteo},
  journal={arXiv preprint arXiv:2501.10573},
  year={2025}
}

@inproceedings{kornblith2019similarity,
  title={Similarity of neural network representations revisited},
  author={Kornblith, Simon and Norouzi, Mohammad and Lee, Honglak and Hinton, Geoffrey},
  booktitle={International conference on machine learning},
  pages={3519--3529},
  year={2019},
  organization={PMlR}
}

@article{huh2024platonic,
  title={The platonic representation hypothesis},
  author={Huh, Minyoung and Cheung, Brian and Wang, Tongzhou and Isola, Phillip},
  journal={arXiv preprint arXiv:2405.07987},
  year={2024}
}

@article{lee2025shared,
  title={Shared global and local geometry of language model embeddings},
  author={Lee, Andrew and Weber, Melanie and Vi{\'e}gas, Fernanda and Wattenberg, Martin},
  journal={arXiv preprint arXiv:2503.21073},
  year={2025}
}

@inproceedings{elliott2024impacts,
  title={The impacts of data, ordering, and intrinsic dimensionality on recall in hierarchical navigable small worlds},
  author={Elliott, Owen P and Clark, Jesse},
  booktitle={Proceedings of the 2024 ACM SIGIR International Conference on Theory of Information Retrieval},
  pages={25--33},
  year={2024}
}

@inproceedings{aumuller2019role,
  title={The role of local intrinsic dimensionality in benchmarking nearest neighbor search},
  author={Aum{\"u}ller, Martin and Ceccarello, Matteo},
  booktitle={International Conference on Similarity Search and Applications},
  pages={113--127},
  year={2019},
  organization={Springer}
}

@article{weller2025theoretical,
  title={On the Theoretical Limitations of Embedding-Based Retrieval},
  author={Weller, Orion and Boratko, Michael and Naim, Iftekhar and Lee, Jinhyuk},
  journal={arXiv preprint arXiv:2508.21038},
  year={2025}
}

@inproceedings{wang2020understanding,
  title={Understanding contrastive representation learning through alignment and uniformity on the hypersphere},
  author={Wang, Tongzhou and Isola, Phillip},
  booktitle={International conference on machine learning},
  pages={9929--9939},
  year={2020},
  organization={PMLR}
}

@inproceedings{wilkie2014best,
  title={Best and fairest: An empirical analysis of retrieval system bias},
  author={Wilkie, Colin and Azzopardi, Leif},
  booktitle={European Conference on Information Retrieval},
  pages={13--25},
  year={2014},
  organization={Springer}
}

@inproceedings{wilkie2014efficiently,
  title={Efficiently estimating retrievability bias},
  author={Wilkie, Colin and Azzopardi, Leif},
  booktitle={European Conference on Information Retrieval},
  pages={720--726},
  year={2014},
  organization={Springer}
}

@inproceedings{adams2020fractal,
  title={A fractal dimension for measures via persistent homology},
  author={Adams, Henry and Aminian, Manuchehr and Farnell, Elin and Kirby, Michael and Mirth, Joshua and Neville, Rachel and Peterson, Chris and Shonkwiler, Clayton},
  booktitle={Topological Data Analysis: The Abel Symposium 2018},
  pages={1--31},
  year={2020},
  organization={Springer}
}

@article{chintakunta2015entropy,
  title={An entropy-based persistence barcode},
  author={Chintakunta, Harish and Gentimis, Thanos and Gonzalez-Diaz, Rocio and Jimenez, Maria-Jose and Krim, Hamid},
  journal={Pattern Recognition},
  volume={48},
  number={2},
  pages={391--401},
  year={2015},
  publisher={Elsevier}
}

@article{facco2017estimating,
  title={Estimating the intrinsic dimension of datasets by a minimal neighborhood information},
  author={Facco, Elena and d’Errico, Maria and Rodriguez, Alex and Laio, Alessandro},
  journal={Scientific reports},
  volume={7},
  number={1},
  pages={12140},
  year={2017},
  publisher={Nature Publishing Group UK London}
}

@article{fukunaga1971algorithm,
  title={An algorithm for finding intrinsic dimensionality of data},
  author={Fukunaga, Keinosuke and Olsen, David R},
  journal={IEEE Transactions on computers},
  volume={100},
  number={2},
  pages={176--183},
  year={1971},
  publisher={IEEE}
}

@inproceedings{basu1996bounding,
  title={On bounding the Betti numbers and computing the Euler characteristic of semi-algebraic sets},
  author={Basu, Saugata},
  booktitle={Proceedings of the twenty-eighth annual ACM symposium on Theory of computing},
  pages={408--417},
  year={1996}
}

@inproceedings{roy2007effective,
  title={The effective rank: A measure of effective dimensionality},
  author={Roy, Olivier and Vetterli, Martin},
  booktitle={2007 15th European signal processing conference},
  pages={606--610},
  year={2007},
  organization={IEEE}
}

@article{leinster2013asymptotic,
  title={On the asymptotic magnitude of subsets of Euclidean space},
  author={Leinster, Tom and Willerton, Simon},
  journal={Geometriae Dedicata},
  volume={164},
  number={1},
  pages={287--310},
  year={2013},
  publisher={Springer}
}

@article{willerton2015spread,
  title={Spread: a measure of the size of metric spaces},
  author={Willerton, Simon},
  journal={International Journal of Computational Geometry \& Applications},
  volume={25},
  number={03},
  pages={207--225},
  year={2015},
  publisher={World Scientific}
}

@article{friedman2022vendi,
  title={The vendi score: A diversity evaluation metric for machine learning},
  author={Friedman, Dan and Dieng, Adji Bousso},
  journal={arXiv preprint arXiv:2210.02410},
  year={2022}
}

@article{rudman2021isoscore,
  title={IsoScore: Measuring the uniformity of embedding space utilization},
  author={Rudman, William and Gillman, Nate and Rayne, Taylor and Eickhoff, Carsten},
  journal={arXiv preprint arXiv:2108.07344},
  year={2021}
}

@inproceedings{shahapure2020cluster,
  title={Cluster quality analysis using silhouette score},
  author={Shahapure, Ketan Rajshekhar and Nicholas, Charles},
  booktitle={2020 IEEE 7th international conference on data science and advanced analytics (DSAA)},
  pages={747--748},
  year={2020},
  organization={IEEE}
}

@article{nguyen2016ms,
  title={Ms marco: A human-generated machine reading comprehension dataset},
  author={Nguyen, Tri and Rosenberg, Mir and Song, Xia and Gao, Jianfeng and Tiwary, Saurabh and Majumder, Rangan and Deng, Li},
  year={2016}
}

@inproceedings{thakur2021beir,
title={{BEIR}: A Heterogeneous Benchmark for Zero-shot Evaluation of Information Retrieval Models},
author={Nandan Thakur and Nils Reimers and Andreas R{\"u}ckl{\'e} and Abhishek Srivastava and Iryna Gurevych},
booktitle={Thirty-fifth Conference on Neural Information Processing Systems Datasets and Benchmarks Track (Round 2)},
year={2021},
url={https://openreview.net/forum?id=wCu6T5xFjeJ}
}

@article{maia2018financial,
  title={Financial opinion mining and question answering},
  author={Maia, M and Handschuh, S and Freitas, A and Davis, B and McDermott, R and Zarrouk, M and others},
  journal={URL: https://sites. google. com/view/fiqa/home},
  year={2018}
}

@misc{diggelmann2021climatefever,
  archiveprefix = {arXiv},
  author = {Thomas Diggelmann and Jordan Boyd-Graber and Jannis Bulian and Massimiliano Ciaramita and Markus Leippold},
  eprint = {2012.00614},
  primaryclass = {cs.CL},
  title = {CLIMATE-FEVER: A Dataset for Verification of Real-World Climate Claims},
  year = {2021},
}

@inproceedings{boteva2016,
  author = {Boteva, Vera and Gholipour, Demian and Sokolov, Artem and Riezler, Stefan},
  city = {Padova},
  country = {Italy},
  journal = {Proceedings of the 38th European Conference on Information Retrieval},
  journal-abbrev = {ECIR},
  title = {A Full-Text Learning to Rank Dataset for Medical Information Retrieval},
  url = {http://www.cl.uni-heidelberg.de/~riezler/publications/papers/ECIR2016.pdf},
  year = {2016},
}

@inproceedings{wachsmuth2018retrieval,
  title={Retrieval of the best counterargument without prior topic knowledge},
  author={Wachsmuth, Henning and Syed, Shahbaz and Stein, Benno},
  booktitle={Proceedings of the 56th Annual Meeting of the Association for Computational Linguistics (Volume 1: Long Papers)},
  pages={241--251},
  year={2018}
}

@misc{voorhees2020treccovidconstructingpandemicinformation,
      title={TREC-COVID: Constructing a Pandemic Information Retrieval Test Collection}, 
      author={Ellen Voorhees and Tasmeer Alam and Steven Bedrick and Dina Demner-Fushman and William R Hersh and Kyle Lo and Kirk Roberts and Ian Soboroff and Lucy Lu Wang},
      year={2020},
      eprint={2005.04474},
      archivePrefix={arXiv},
      primaryClass={cs.IR},
      url={https://arxiv.org/abs/2005.04474}, 
}

@misc{wadden2020factfictionverifyingscientific,
      title={Fact or Fiction: Verifying Scientific Claims}, 
      author={David Wadden and Shanchuan Lin and Kyle Lo and Lucy Lu Wang and Madeleine van Zuylen and Arman Cohan and Hannaneh Hajishirzi},
      year={2020},
      eprint={2004.14974},
      archivePrefix={arXiv},
      primaryClass={cs.CL},
      url={https://arxiv.org/abs/2004.14974}, 
}

@misc{cohan2020specterdocumentlevelrepresentationlearning,
      title={SPECTER: Document-level Representation Learning using Citation-informed Transformers}, 
      author={Arman Cohan and Sergey Feldman and Iz Beltagy and Doug Downey and Daniel S. Weld},
      year={2020},
      eprint={2004.07180},
      archivePrefix={arXiv},
      primaryClass={cs.CL},
      url={https://arxiv.org/abs/2004.07180}, 
}

@inproceedings{hoogeveen2015cqadupstack,
  title={Cqadupstack: A benchmark data set for community question-answering research},
  author={Hoogeveen, Doris and Verspoor, Karin M and Baldwin, Timothy},
  booktitle={Proceedings of the 20th Australasian document computing symposium},
  pages={1--8},
  year={2015}
}

@article{edelsbrunner2008persistent,
  title={Persistent homology-a survey},
  author={Edelsbrunner, Herbert and Harer, John and others},
  journal={Contemporary mathematics},
  volume={453},
  number={26},
  pages={257--282},
  year={2008},
  publisher={Providence, RI: American Mathematical Society}
}

@article{li2023towards,
  title={Towards general text embeddings with multi-stage contrastive learning},
  author={Li, Zehan and Zhang, Xin and Zhang, Yanzhao and Long, Dingkun and Xie, Pengjun and Zhang, Meishan},
  journal={arXiv preprint arXiv:2308.03281},
  year={2023}
}

@misc{zhang2025jasperstelladistillationsota,
      title={Jasper and Stella: distillation of SOTA embedding models}, 
      author={Dun Zhang and Jiacheng Li and Ziyang Zeng and Fulong Wang},
      year={2025},
      eprint={2412.19048},
      archivePrefix={arXiv},
      primaryClass={cs.IR},
      url={https://arxiv.org/abs/2412.19048}, 
}

@article{qwen3embedding,
  title={Qwen3 Embedding: Advancing Text Embedding and Reranking Through Foundation Models},
  author={Zhang, Yanzhao and Li, Mingxin and Long, Dingkun and Zhang, Xin and Lin, Huan and Yang, Baosong and Xie, Pengjun and Yang, An and Liu, Dayiheng and Lin, Junyang and Huang, Fei and Zhou, Jingren},
  journal={arXiv preprint arXiv:2506.05176},
  year={2025}
}

@misc{LinqAIResearch2024,
  title = {Linq-Embed-Mistral: Elevating Text Retrieval with Improved GPT Data Through Task-Specific Control and Quality Refinement},
  author = {Junseong Kim and Seolhwa Lee and Jihoon Kwon and Sangmo Gu and Yejin Kim and Minkyung Cho and Jy-yong Sohn and Chanyeol Choi},
  year = {2024},
  howpublished = {\url{https://getlinq.com/blog/linq-embed-mistral/}},
  note = {Linq AI Research Blog}
}

@misc{SFRAIResearch2024,
  title = {SFR-Embedding-Mistral: Enhance Text Retrieval with Transfer Learning},
  author = {Rui Meng and Ye Liu and Shafiq Rayhan Joty and Caiming Xiong and Yingbo Zhou and Semih Yavuz},
  year = {2024},
  howpublished = {\url{https://www.salesforce.com/blog/sfr-embedding/}},
  note = {Salesforce AI Research Blog}
}

@article{wang2022text,
  title={Text Embeddings by Weakly-Supervised Contrastive Pre-training},
  author={Wang, Liang and Yang, Nan and Huang, Xiaolong and Jiao, Binxing and Yang, Linjun and Jiang, Daxin and Majumder, Rangan and Wei, Furu},
  journal={arXiv preprint arXiv:2212.03533},
  year={2022}
}

@misc{bge_embedding,
      title={C-Pack: Packaged Resources To Advance General Chinese Embedding}, 
      author={Shitao Xiao and Zheng Liu and Peitian Zhang and Niklas Muennighoff},
      year={2023},
      eprint={2309.07597},
      archivePrefix={arXiv},
      primaryClass={cs.CL}
}

@article{wang2024multilingual,
  title={Multilingual E5 Text Embeddings: A Technical Report},
  author={Wang, Liang and Yang, Nan and Huang, Xiaolong and Yang, Linjun and Majumder, Rangan and Wei, Furu},
  journal={arXiv preprint arXiv:2402.05672},
  year={2024}
}

@article{merrick2024arctic,
  title={Arctic-embed: Scalable, efficient, and accurate text embedding models},
  author={Merrick, Luke and Xu, Danmei and Nuti, Gaurav and Campos, Daniel},
  journal={arXiv preprint arXiv:2405.05374},
  year={2024}
}

@misc{nur2024emberv1,
      title={ember-v1: SOTA embedding model}, 
      author={Enrike Nur and Anar Aliyev},
      year={2023},
}

@online{xsmall2024mxbai,
  title={Every Byte Matters: Introducing mxbai-embed-xsmall-v1},
  author={Sean Lee and Julius Lipp and Rui Huang and Darius Koenig},
  year={2024},
  url={https://www.mixedbread.ai/blog/mxbai-embed-xsmall-v1},
}

@online{emb2024mxbai,
  title={Open Source Strikes Bread - New Fluffy Embeddings Model},
  author={Sean Lee and Aamir Shakir and Darius Koenig and Julius Lipp},
  year={2024},
  url={https://www.mixedbread.ai/blog/mxbai-embed-large-v1},
}

@misc{wang2020minilmdeepselfattentiondistillation,
      title={MiniLM: Deep Self-Attention Distillation for Task-Agnostic Compression of Pre-Trained Transformers}, 
      author={Wenhui Wang and Furu Wei and Li Dong and Hangbo Bao and Nan Yang and Ming Zhou},
      year={2020},
      eprint={2002.10957},
      archivePrefix={arXiv},
      primaryClass={cs.CL},
      url={https://arxiv.org/abs/2002.10957}, 
}

@misc{ni2021largedualencodersgeneralizable,
      title={Large Dual Encoders Are Generalizable Retrievers}, 
      author={Jianmo Ni and Chen Qu and Jing Lu and Zhuyun Dai and Gustavo Hernández Ábrego and Ji Ma and Vincent Y. Zhao and Yi Luan and Keith B. Hall and Ming-Wei Chang and Yinfei Yang},
      year={2021},
      eprint={2112.07899},
      archivePrefix={arXiv},
      primaryClass={cs.IR},
      url={https://arxiv.org/abs/2112.07899}, 
}

@article{xiong2020approximate,
  title={Approximate nearest neighbor negative contrastive learning for dense text retrieval},
  author={Xiong, Lee and Xiong, Chenyan and Li, Ye and Tang, Kwok-Fung and Liu, Jialin and Bennett, Paul and Ahmed, Junaid and Overwijk, Arnold},
  journal={arXiv preprint arXiv:2007.00808},
  year={2020}
}

@inproceedings{Hofstaetter2021_tasb_dense_retrieval,
 author = {Sebastian Hofst{\"a}tter and Sheng-Chieh Lin and Jheng-Hong Yang and Jimmy Lin and Allan Hanbury},
 title = {{Efficiently Teaching an Effective Dense Retriever with Balanced Topic Aware Sampling}},
 booktitle = {Proc. of SIGIR},
 year = {2021},
}

@misc{lee2025geminiembeddinggeneralizableembeddings,
      title={Gemini Embedding: Generalizable Embeddings from Gemini}, 
      author={Jinhyuk Lee and Feiyang Chen and Sahil Dua and Daniel Cer and Madhuri Shanbhogue and Iftekhar Naim and Gustavo Hernández Ábrego and Zhe Li and Kaifeng Chen and Henrique Schechter Vera and Xiaoqi Ren and Shanfeng Zhang and Daniel Salz and Michael Boratko and Jay Han and Blair Chen and Shuo Huang and Vikram Rao and Paul Suganthan and Feng Han and Andreas Doumanoglou and Nithi Gupta and Fedor Moiseev and Cathy Yip and Aashi Jain and Simon Baumgartner and Shahrokh Shahi and Frank Palma Gomez and Sandeep Mariserla and Min Choi and Parashar Shah and Sonam Goenka and Ke Chen and Ye Xia and Koert Chen and Sai Meher Karthik Duddu and Yichang Chen and Trevor Walker and Wenlei Zhou and Rakesh Ghiya and Zach Gleicher and Karan Gill and Zhe Dong and Mojtaba Seyedhosseini and Yunhsuan Sung and Raphael Hoffmann and Tom Duerig},
      year={2025},
      eprint={2503.07891},
      archivePrefix={arXiv},
      primaryClass={cs.CL},
      url={https://arxiv.org/abs/2503.07891}, 
}

@misc{muennighoff2023mtebmassivetextembedding,
      title={MTEB: Massive Text Embedding Benchmark}, 
      author={Niklas Muennighoff and Nouamane Tazi and Loïc Magne and Nils Reimers},
      year={2023},
      eprint={2210.07316},
      archivePrefix={arXiv},
      primaryClass={cs.CL},
      url={https://arxiv.org/abs/2210.07316}, 
}

@article{lewis2020retrieval,
  title={Retrieval-augmented generation for knowledge-intensive nlp tasks},
  author={Lewis, Patrick and Perez, Ethan and Piktus, Aleksandra and Petroni, Fabio and Karpukhin, Vladimir and Goyal, Naman and K{\"u}ttler, Heinrich and Lewis, Mike and Yih, Wen-tau and Rockt{\"a}schel, Tim and others},
  journal={Advances in neural information processing systems},
  volume={33},
  pages={9459--9474},
  year={2020}
}

@article{petukhova2025text,
  title={Text clustering with large language model embeddings},
  author={Petukhova, Alina and Matos-Carvalho, Joao P and Fachada, Nuno},
  journal={International Journal of Cognitive Computing in Engineering},
  volume={6},
  pages={100--108},
  year={2025},
  publisher={Elsevier}
}

@article{da2023text,
  title={Text classification using embeddings: a survey},
  author={Da Costa, Liliane Soares and Oliveira, Italo L and Fileto, Renato},
  journal={Knowledge and Information Systems},
  volume={65},
  number={7},
  pages={2761--2803},
  year={2023},
  publisher={Springer}
}

@article{zhao2024dense,
  title={Dense text retrieval based on pretrained language models: A survey},
  author={Zhao, Wayne Xin and Liu, Jing and Ren, Ruiyang and Wen, Ji-Rong},
  journal={ACM Transactions on Information Systems},
  volume={42},
  number={4},
  pages={1--60},
  year={2024},
  publisher={ACM New York, NY}
}

@article{darrin2024embedding,
  title={When is an Embedding Model More Promising than Another?},
  author={Darrin, Maxime and Formont, Philippe and Ayed, Ismail and Cheung, Jackie CK and Piantanida, Pablo},
  journal={Advances in Neural Information Processing Systems},
  volume={37},
  pages={68330--68379},
  year={2024}
}

@article{caspari2024beyond,
  title={Beyond benchmarks: Evaluating embedding model similarity for retrieval augmented generation systems},
  author={Caspari, Laura and Dastidar, Kanishka Ghosh and Zerhoudi, Saber and Mitrovic, Jelena and Granitzer, Michael},
  journal={arXiv preprint arXiv:2407.08275},
  year={2024}
}

@inproceedings{devlin2019bert,
  title={Bert: Pre-training of deep bidirectional transformers for language understanding},
  author={Devlin, Jacob and Chang, Ming-Wei and Lee, Kenton and Toutanova, Kristina},
  booktitle={Proceedings of the 2019 conference of the North American chapter of the association for computational linguistics: human language technologies, volume 1 (long and short papers)},
  pages={4171--4186},
  year={2019}
}

@article{wang2019evaluating,
  title={Evaluating word embedding models: Methods and experimental results},
  author={Wang, Bin and Wang, Angela and Chen, Fenxiao and Wang, Yuncheng and Kuo, C-C Jay},
  journal={APSIPA transactions on signal and information processing},
  volume={8},
  pages={e19},
  year={2019},
  publisher={Cambridge University Press}
}

@article{lee2024geometric,
  title={Geometric Signatures of Compositionality Across a Language Model's Lifetime},
  author={Lee, Jin Hwa and Jiralerspong, Thomas and Yu, Lei and Bengio, Yoshua and Cheng, Emily},
  journal={arXiv preprint arXiv:2410.01444},
  year={2024}
}

@article{sucholutsky2023aligned,
  title={Getting aligned on representational alignment},
  author={Sucholutsky, Ilia and Muttenthaler, Lukas and Weller, Adrian and Peng, Andi and Bobu, Andreea and Kim, Been and Love, Bradley C and Grant, Erin and Groen, Iris and Achterberg, Jascha and others},
  journal={arXiv preprint arXiv:2310.13018},
  year={2023}
}

@article{robertson2009probabilistic,
  title={The probabilistic relevance framework: BM25 and beyond},
  author={Robertson, Stephen and Zaragoza, Hugo and others},
  journal={Foundations and Trends{\textregistered} in Information Retrieval},
  volume={3},
  number={4},
  pages={333--389},
  year={2009},
  publisher={Now Publishers, Inc.}
}

@article{pearson1901pca,
  title={On lines and planes of closest fit to systems of points in space},
  author={Pearson, Karl},
  journal={Philosophical Magazine},
  volume={2},
  number={11},
  pages={559--572},
  year={1901},
  publisher={Taylor \& Francis}
}

@misc{wolf2020huggingfacestransformersstateoftheartnatural,
      title={HuggingFace's Transformers: State-of-the-art Natural Language Processing}, 
      author={Thomas Wolf and Lysandre Debut and Victor Sanh and Julien Chaumond and Clement Delangue and Anthony Moi and Pierric Cistac and Tim Rault and Rémi Louf and Morgan Funtowicz and Joe Davison and Sam Shleifer and Patrick von Platen and Clara Ma and Yacine Jernite and Julien Plu and Canwen Xu and Teven Le Scao and Sylvain Gugger and Mariama Drame and Quentin Lhoest and Alexander M. Rush},
      year={2020},
      eprint={1910.03771},
      archivePrefix={arXiv},
      primaryClass={cs.CL},
      url={https://arxiv.org/abs/1910.03771}, 
}

@incollection{kramer2016scikit,
  title={Scikit-learn},
  author={Kramer, Oliver},
  booktitle={Machine learning for evolution strategies},
  pages={45--53},
  year={2016},
  publisher={Springer}
}

@article{gupta2024geometric,
  title={Geometric Interpretation of Layer Normalization and a Comparative Analysis with RMSNorm},
  author={Gupta, Akshat and Ozdemir, Atahan and Anumanchipalli, Gopala},
  journal={arXiv preprint arXiv:2409.12951},
  year={2024}
}

@inproceedings{wilkie2013relating,
  title={Relating retrievability, performance and length},
  author={Wilkie, Colin and Azzopardi, Leif},
  booktitle={Proceedings of the 36th international ACM SIGIR conference on Research and development in information retrieval},
  pages={937--940},
  year={2013}
}

@article{mickus2024isotropy,
   title={Isotropy, clusters, and classifiers},
   author={Mickus, Timothee and Gr{\"o}nroos, Stig-Arne and Attieh, Joseph},
   journal={arXiv preprint arXiv:2402.03191},
   year={2024}
 }

@article{mcinnes2018umap,
  title={Umap: Uniform manifold approximation and projection for dimension reduction},
  author={McInnes, Leland and Healy, John and Melville, James},
  journal={arXiv preprint arXiv:1802.03426},
  year={2018}
}

@misc{tao2025llmseffectiveembeddingmodels,
      title={LLMs are Also Effective Embedding Models: An In-depth Overview}, 
      author={Chongyang Tao and Tao Shen and Shen Gao and Junshuo Zhang and Zhen Li and Kai Hua and Wenpeng Hu and Zhengwei Tao and Shuai Ma},
      year={2025},
      eprint={2412.12591},
      archivePrefix={arXiv},
      primaryClass={cs.CL},
      url={https://arxiv.org/abs/2412.12591}, 
}

@inproceedings{ciernikobjective,
  title={Objective drives the consistency of representational similarity across datasets},
  author={Ciernik, Laure and Linhardt, Lorenz and Morik, Marco and Dippel, Jonas and Kornblith, Simon and Muttenthaler, Lukas},
  booktitle={Forty-second International Conference on Machine Learning}
}

@article{ackerman2016characterization,
  title={A characterization of linkage-based hierarchical clustering},
  author={Ackerman, Margareta and Ben-David, Shai},
  journal={Journal of Machine Learning Research},
  volume={17},
  number={231},
  pages={1--17},
  year={2016}
}

@misc{gruver2024largelanguagemodelszeroshot,
      title={Large Language Models Are Zero-Shot Time Series Forecasters}, 
      author={Nate Gruver and Marc Finzi and Shikai Qiu and Andrew Gordon Wilson},
      year={2024},
      eprint={2310.07820},
      archivePrefix={arXiv},
      primaryClass={cs.LG},
      url={https://arxiv.org/abs/2310.07820}, 
}

@misc{islam2025manifoldapproximationleadsrobust,
      title={Manifold Approximation leads to Robust Kernel Alignment}, 
      author={Mohammad Tariqul Islam and Du Liu and Deblina Sarkar},
      year={2025},
      eprint={2510.22953},
      archivePrefix={arXiv},
      primaryClass={cs.LG},
      url={https://arxiv.org/abs/2510.22953}, 
}

@misc{barannikov2022representationtopologydivergencemethod,
      title={Representation Topology Divergence: A Method for Comparing Neural Network Representations}, 
      author={Serguei Barannikov and Ilya Trofimov and Nikita Balabin and Evgeny Burnaev},
      year={2022},
      eprint={2201.00058},
      archivePrefix={arXiv},
      primaryClass={cs.LG},
      url={https://arxiv.org/abs/2201.00058}, 
}

@article{davari2022reliability,
  title={Reliability of cka as a similarity measure in deep learning},
  author={Davari, MohammadReza and Horoi, Stefan and Natik, Amine and Lajoie, Guillaume and Wolf, Guy and Belilovsky, Eugene},
  journal={arXiv preprint arXiv:2210.16156},
  year={2022}
}

@article{setty2024improvingretrievalragbased,
  title={Improving retrieval for rag based question answering models on financial documents},
  author={Setty, Spurthi and Thakkar, Harsh and Lee, Alyssa and Chung, Eden and Vidra, Natan},
  journal={arXiv preprint arXiv:2404.07221},
  year={2024}
}

@misc{su2025parametricretrievalaugmentedgeneration,
      title={Parametric Retrieval Augmented Generation}, 
      author={Weihang Su and Yichen Tang and Qingyao Ai and Junxi Yan and Changyue Wang and Hongning Wang and Ziyi Ye and Yujia Zhou and Yiqun Liu},
      year={2025},
      eprint={2501.15915},
      archivePrefix={arXiv},
      primaryClass={cs.CL},
      url={https://arxiv.org/abs/2501.15915}, 
}

@misc{limbeck2025metricspacemagnitudeevaluating,
      title={Metric Space Magnitude for Evaluating the Diversity of Latent Representations}, 
      author={Katharina Limbeck and Rayna Andreeva and Rik Sarkar and Bastian Rieck},
      year={2025},
      eprint={2311.16054},
      archivePrefix={arXiv},
      primaryClass={cs.LG},
      url={https://arxiv.org/abs/2311.16054}, 
}

@article{ruppik2025less,
  title={Less is More: Local Intrinsic Dimensions of Contextual Language Models},
  author={Ruppik, Benjamin Matthias and von Rohrscheidt, Julius and van Niekerk, Carel and Heck, Michael and Vukovic, Renato and Feng, Shutong and Lin, Hsien-chin and Lubis, Nurul and Rieck, Bastian and Zibrowius, Marcus and others},
  journal={arXiv preprint arXiv:2506.01034},
  year={2025}
}

@misc{openai2023embedding,
  author       = {OpenAI},
  title        = {New embedding models and API updates},
  year         = {2023},
  howpublished = {\url{https://openai.com/index/new-embedding-models-and-api-updates/}},
  note         = {Accessed: 2025-11-10}
}

@misc{magnipy,
  author       = {Katharina Limbeck and Emily Simons},
  title        = {Magnipy: A Python library for computing magnitude of metric spaces},
  year         = {2025},
  howpublished = {\url{https://github.com/aidos-lab/magnipy}},
  note         = {GitHub repository}
}

@article{bac2021scikit,
  title={Scikit-dimension: a python package for intrinsic dimension estimation},
  author={Bac, Jonathan and Mirkes, Evgeny M and Gorban, Alexander N and Tyukin, Ivan and Zinovyev, Andrei},
  journal={Entropy},
  volume={23},
  number={10},
  pages={1368},
  year={2021},
  publisher={MDPI}
}

@article{Bauer2021Ripser,
    AUTHOR = {Bauer, Ulrich},
     TITLE = {Ripser: efficient computation of {V}ietoris-{R}ips persistence
              barcodes},
   JOURNAL = {J. Appl. Comput. Topol.},
  FJOURNAL = {Journal of Applied and Computational Topology},
    VOLUME = {5},
      YEAR = {2021},
    NUMBER = {3},
     PAGES = {391--423},
      ISSN = {2367-1726},
   MRCLASS = {55N31 (55-04)},
  MRNUMBER = {4298669},
       DOI = {10.1007/s41468-021-00071-5},
       URL = {https://doi.org/10.1007/s41468-021-00071-5},
}

@inproceedings{reimers2021curse,
  title={The curse of dense low-dimensional information retrieval for large index sizes},
  author={Reimers, Nils and Gurevych, Iryna},
  booktitle={Proceedings of the 59th Annual Meeting of the Association for Computational Linguistics and the 11th International Joint Conference on Natural Language Processing (Volume 2: Short Papers)},
  pages={605--611},
  year={2021}
}

@inproceedings{groger2026revisiting,
  title     = {Revisiting the Platonic Representation Hypothesis: An Aristotelian View},
  author    = {Gr{\"o}ger, Fabian and Wen, Shuo and Brbi{\'c}, Maria},
  booktitle = {Proceedings of the 43rd International Conference on Machine Learning (ICML)},
  year      = {2026},
}
\newpage
\appendix
\onecolumn

\section{Computation of individual topological descriptors}
\label{a_desc}

\subsection{Persistent Homology Statistics}
Let $ \{(b_i, d_i)\}^N_{i=1}$ be the birth-death pairs of a Vietoris Rips filtration. We summarize the persistence diagram using:

\[
\text{mean lifetime}=\frac{1}{N}\sum_{i}(d_i-b_i),\qquad
\text{mean midlife}=\frac{1}{N}\sum_{i}\frac{b_i+d_i}{2}\qquad
\]

We use the Python library ripser \cite{Bauer2021Ripser} for all Persistent Homology computations.
\subsection{Persistence Entropy}
Let \(\ell_i=d_i-b_i\) be the lifetimes and \(p_i=\ell_i/\sum_j \ell_j\) the normalized lifetimes. We quantify the complexity of the persistence diagram using the Shannon entropy of the normalized lifetimes:

\[
H=-\sum_i p_i\log p_i.
\]

\subsection{Persistent Homology Dimension}
Let $X$ be a bounded subset of a metric space and define the $\alpha$-weighted sum of the length of persistence intervals, given by their birth $b$ and death $d$ times as:
\begin{equation*}
	E_i^\alpha =\sum_{(b,d) \in PH_i(x)} (d-b)^\alpha
	\label{intervalsum}
\end{equation*}

Then  $dim_i^{PH}$ for the Vietoris-Rips complex of $X$ corresponds to the smallest $\alpha$ for which $E_i^\alpha$ is uniformly bounded for all subsets of $X$:
\begin{equation*}
	dim_i^{PH}(X) = \inf \{\alpha: \exists C \quad s.t. \quad E^i_\alpha(x) < C \\
	\quad \forall x \subset X\}
	\label{phdim}
\end{equation*}

Intuitively, $dim_i^{PH}$ is determined by how the lengths of the PH intervals for increasing sample sizes vary for a specific homological dimension $i$. Higher values suggest a more complex topology across scales, as the total persistence (sum of lifetimes) grows rapidly with sample size.

\subsection{Euler Characteristic}
Let $\beta_i$ be the $i^{th}$ Betti number, counting the number of independent $i$-dimensional holes (connected components, loops, voids $\dots$) in $X$. Then the Euler Characteristic is given by alternating sums of $\beta_i$:

\[
\chi=\sum_{i=0}^n (-1)^i \beta_i,
\]

\subsection{TwoNN Dimension}
Let \( d_{1,i} \) and \( d_{2,i} \) be the distances from point \( x_i \) to its first and second nearest neighbors and define the ratio $\mu_i = \frac{d_{2,i}}{d_{1,i}}$. The intrinsic dimension $d$ is computed based on the following equation:

\[
-\log(1 - F(\mu)) = d \cdot \log(\mu),
\]

where $F(\mu)$ is the empirical cumulative distribution function of $\mu$. Fitting the points $(\log \mu, -\log(1-F(\mu)))$ with a line through the origin yields the estimate of $d$. We use the Python library scikit-dimension \cite{bac2021scikit} for computations.

\subsection{PCA Dimension}
Let $\lambda_1 \geq \lambda_2 \geq \dots \geq \lambda_d$ be the eigenvalues of the covariance matrix of the dataset.  
The Fukunaga--Olsen (FO) rule estimates the intrinsic dimension as the number of eigenvalues larger than a fraction $\alpha_{\text{FO}}$ of the largest eigenvalue:

\[
\hat{m}_{\text{FO}} = \# \left\{ \lambda_j \;\middle|\; \lambda_j \geq \alpha_{\text{FO}} \cdot \lambda_{\max} \right\},
\]

where $\lambda_{\max} = \max_j \lambda_j$. We use the Python library scikit-dimension \cite{bac2021scikit} for computations with $\alpha_{\text{FO}} = 0.5$.

\subsection{Effective Rank}
Given normalized eigenvalues \(p_j=\lambda_j/\sum_k\lambda_k\) of the covariance matrix of $X$, the effective rank is defined as the Shannon entropy, reflecting the effective number of principal components:

\[
\mathrm{ER}=\exp\Big(-\sum_j p_j\log p_j\Big).
\]

\subsection{Magnitude and Magnitude function}
Given a finite metric space $X = \{x_1, \ldots, x_n\} \subseteq \mathbb{R}^D$ and the corresponding similarity matrix $\zeta_X(i,j) := \exp\!\bigl(-d(x_i, x_j)\bigr)$, then for invertible $\zeta_X(i,j)$ the magnitude of X is defined as:

\[
\mathrm{Mag}(X) := \sum_{i,j} \bigl(\zeta_X^{-1}\bigr)_{ij}
\]

Using a scaling factor $t \in \mathbb{R}^+$, magnitude can be computed for different scales $d_t(x,y)= t \cdot d(x,y)$ yielding the magnitude function:

\[
\mathrm{Mag}_X : t \;\mapsto\; \mathrm{Mag}(tX)
\]

\subsection{Magnitude Dimension}
As the scaling factor $t$ grows, the following limit characterizes the asymptotic growth rate of the magnitude function:
\[
\dim_{\mathrm{Mag}}(X) := \lim_{t \to \infty} \frac{\log \mathrm{Mag}(tX)}{\log t},
\]
For the computation we use the Python library magnipy \cite{magnipy}.

\subsection{Magnitude Area}
The area under the magnitude function is defined as:

\[
\mathrm{MagArea}=\int_{t_0}^{t_{cut}}\mathrm{Mag_X}(t)\,dt.
\]
For the computation we use the Python library magnipy \cite{magnipy}.

\subsection{Spread}
Given a finite metric space \(X\) with metric \(d\), the spread \(E_0(X)\) is measuring the size of the metric space by:

\[
E_0(X) := \sum_{x \in X} \frac{1}{\sum_{x' \in X} e^{-d(x,x')}}.
\]

\subsection{Vendi Score}
Let $X = \{x_1, \ldots, x_n\} \subseteq \mathbb{R}^D$ be a collection of samples and $K \in \mathbb{R}^{n \times n}$ the corresponding similarity (or more generally kernel) matrix. The Vendi Score is defined as the Shannon entropy of the eigenvalues $\lambda_1, \dots \lambda_k$ of $K/n$:

\[
\mathrm{VS}_k(x_1, \dots x_n)=\exp\Big(-\sum_i \lambda_i\log \lambda_i\Big).
\]

\subsection{Mean Pairwise Similarity}
Density-like descriptor computed from pairwise similarities used as average similarity estimate. We use Cosine similarity and Euclidean similarity as kernels \(k\) and compute

\[
\bar{s}=\frac{1}{n}\sum_{i}\frac{1}{n}\sum_{j} k(x_i,x_j).
\]

\subsection{Uniformity}
Let $X = \{x_1, \ldots, x_n\} \subseteq \mathbb{R}^D$ be a collection of samples and $G_t: X^d \times X^d \rightarrow \mathbb{R}_{+}$ the Radial Basis Function kernel, defined as:
\[
G_t(u,v) \triangleq e^{-t\|u-v\|^2_2} = e^{2t \cdot u^\mathsf{T}v - 2t}, \quad t>0,
\]

The uniformity metric \cite{wang2020understanding} is defined as the logarithm of the average pairwise kernel:
\[
\mathcal{L}_{\text{uniform}}(f;t) \triangleq \log \underset{x,y \sim p_{\text{data}}}{\mathbb{E}_{\text{i.i.d.}}} [G_t(u,v)] = \log \underset{x,y \sim p_{\text{data}}}{\mathbb{E}_{\text{i.i.d.}}} \left[e^{-t\|f(x)-f(y)\|^2_2}\right], \quad t>0.
\]

\subsection{IsoScore}
Given a point cloud $X = \{x_1, \ldots, x_n\} \subseteq \mathbb{R}^D$ and its transformation onto the first $k$ principal components $X^{PCA}$. Define the eigenvalues $\hat{\Sigma_D}$ as the normalized diagonal of the covariance matrix of $X^{PCA}$. Then IsoScore can be summarized using the following equation:
\[
\text{IsoScore}(X) = \frac{n \left( \left( n - ||\hat{\Sigma}_D - \mathbf{1}|| / \sqrt{2(n - \sqrt{n})} \right)^2 (n - \sqrt{n})^2 / n^2 \right) - 1}{n - 1}
\]

\subsection{Silhouette Score}
For a set of points $X = \{x_1, \ldots, x_n\} \subseteq \mathbb{R}^D$ we can measure how well points match their assigned clusters. For point \(i\), let \(a_i\) be the mean intra-cluster distance and \(b_i\) the mean distance to the nearest other cluster. The Silhouette Score is defined as:

\[
s_i=\frac{b_i-a_i}{\max(a_i,b_i)},\qquad
\mathrm{Silhouette}=\frac{1}{n}\sum_{i=1}^n s_i.
\]

We apply the \textit{k-means} algorithm to partition the embedding spaces into clusters. 
To identify the optimal number of clusters, we evaluate \[
k \in \{3, 5, 10, 20, 50, 100\}
\] and compute the corresponding Silhouette Scores. 
The value of \(k\) that maximizes the Silhouette Score is selected and subsequently used 
as the metric for characterizing the embedding space.

\section{Scaling analysis}

\label{a_scale}
We perform an empirical scaling analysis across all descriptors to identify an appropriate sample size. This analysis accounts for heterogeneous convergence properties and the varying computational costs associated with the descriptors.

\subsection{Convergence}
First, we investigate the stability of the measures with increasing sample size, to ensure that the signatures are robust. We found that the metrics react very differently to adjusting the number of samples, for example magnitude area is increasing with sample size, while pairwise distances are converging quickly. An example of this analysis for all datasets is provided in Figure \ref*{rank_scaling}, which shows that the measure effective rank starts to converge after 10000 samples. This implies, as some datasets have a lower number of documents, some measures might not fully converge. Additionally, Figure \ref*{rank_scaling} highlights that the absolute values can vary across datasets, while the general convergence pattern is consistent. 

\subsection{Computational complexity}

\begin{figure*}[t] 
    \vskip 0.2in
    \begin{center}
        \begin{minipage}{0.5\textwidth}
            \centerline{\includegraphics[width=\columnwidth]{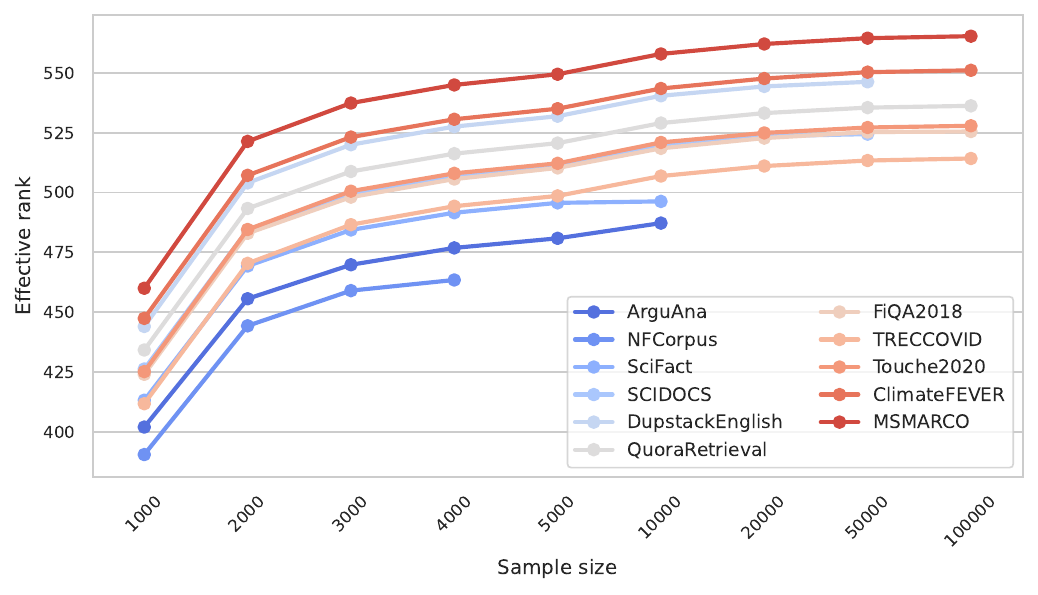}}
            \caption{Convergence of the effective rank.}
            \label{rank_scaling}
        \end{minipage}%
        \begin{minipage}{0.5\textwidth}
            \centerline{\includegraphics[width=\columnwidth]{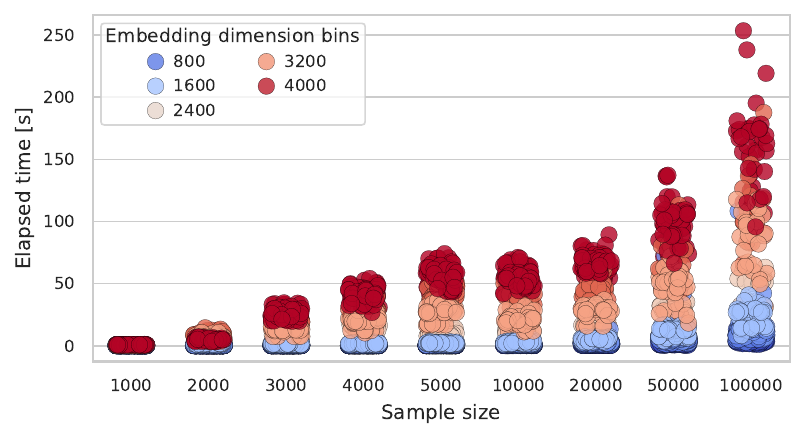}}
            \caption{Time complexity of effective rank.}
            \label{compute_scaling}
        \end{minipage}
    \end{center} 
    \vskip -0.2in 
\end{figure*}

In addition, we also analyzed the computation time in seconds for each sample size, as shown in Figure \ref*{compute_scaling} for effective rank. We observe that the embedding dimension has a strong influence on the computation time for most descriptors, as higher dimensions lead to more expensive computations. Based on these results, we choose the sample sizes as a balance between computational efficiency and descriptor robustness.

\section{Disentangling dataset and model contributions}
\label{a_disentangle}
\begin{figure}[t]
    \vskip 0.2in
    \begin{center}
        \centerline{\includegraphics[width=\columnwidth]{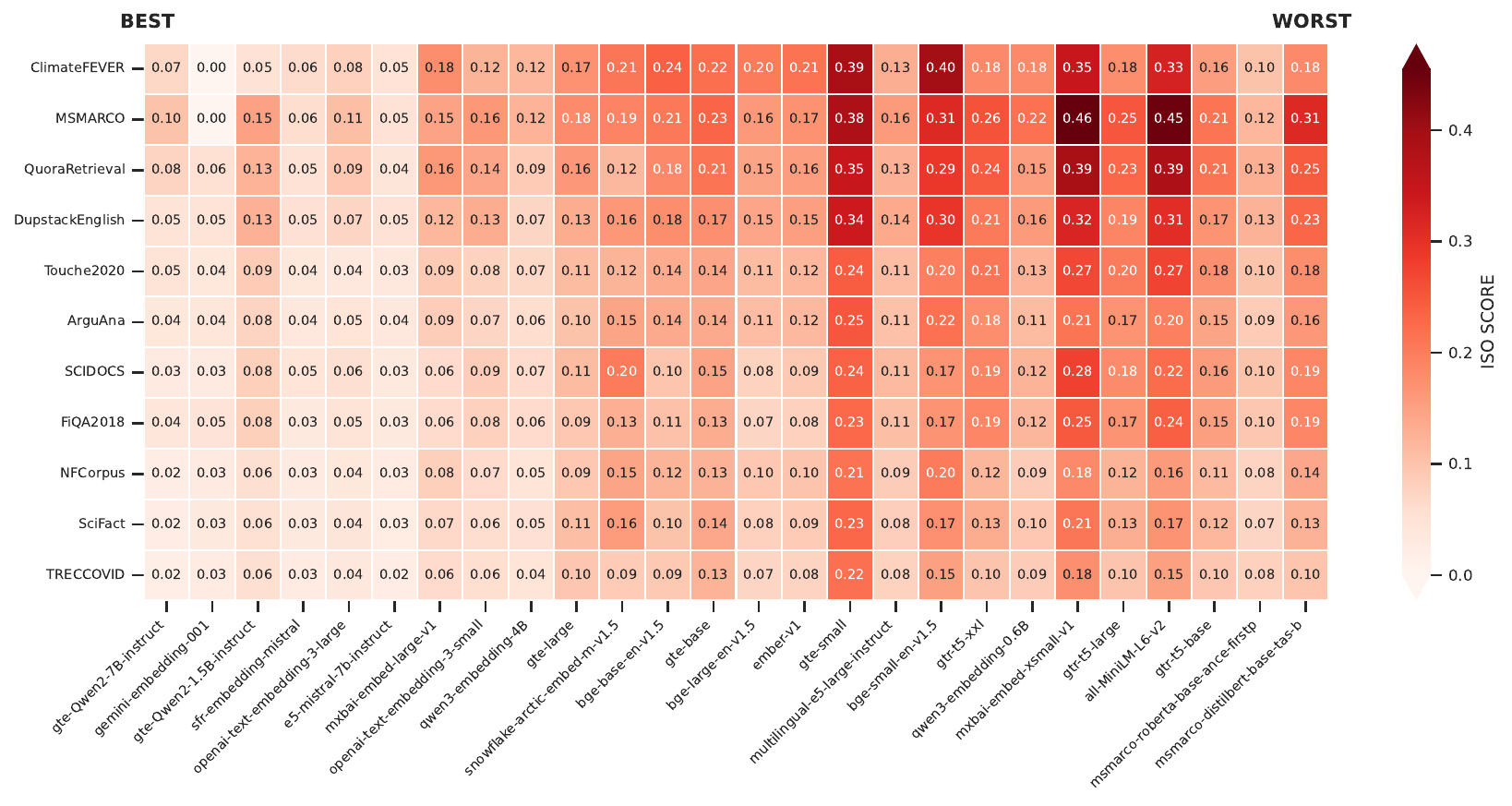}}
        \caption{IsoScore of embedding spaces by different models and datasets sorted by MTEB score.}
        \label{heatmap-iso}
        
        \centerline{\includegraphics[width=\columnwidth]{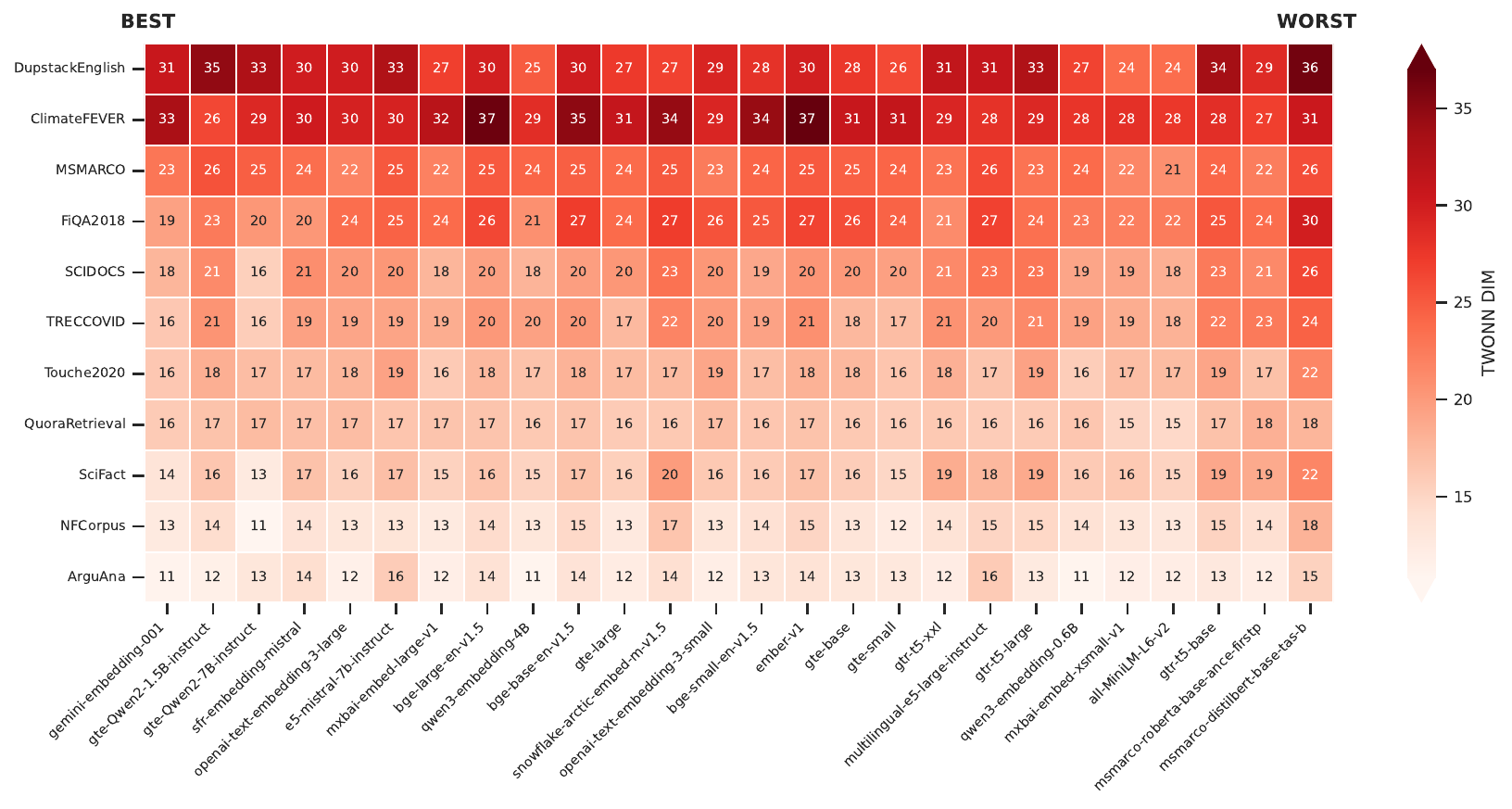}}
        \caption{TwoNN dimension of embedding spaces by different models and datasets sorted by MTEB score.}
        \label{heatmap-twonn}
    \end{center}
    \vskip -0.2in
\end{figure}

We observe a variation in descriptor behavior: some characterize the dataset, while others serve as markers of the embedding model itself. For instance, Figure \ref*{heatmap-iso} shows a clear anisotropy pattern across models, while the values of different datasets show only little variance for a specific model. This indicates, that this descriptor is suitable for distinguishing between models, but not between datasets. In contrast, Figure \ref*{heatmap-twonn} shows that intrinsic dimension measured through TwoNN appears to be a property of datasets rather than models, showing significant variance across datasets. A part of this effect can be explained by larger dataset sizes, resulting in a higher intrinsic complexity. These findings have important implications for our normalization scheme. As we normalize the UTS vectors across datasets, we thereby also remove the variability in some descriptors, causing them to be less discriminative. As we are mostly interested in the effects of embedding models, this choice is consistent with our objective of highlighting model-related properties rather than dataset-driven effects.

\section{Correlation analysis}
\label{a_corr}
\begin{figure*}[t]
    \vskip 0.2in
    \begin{center}
    \centerline{\includegraphics[width=\columnwidth]{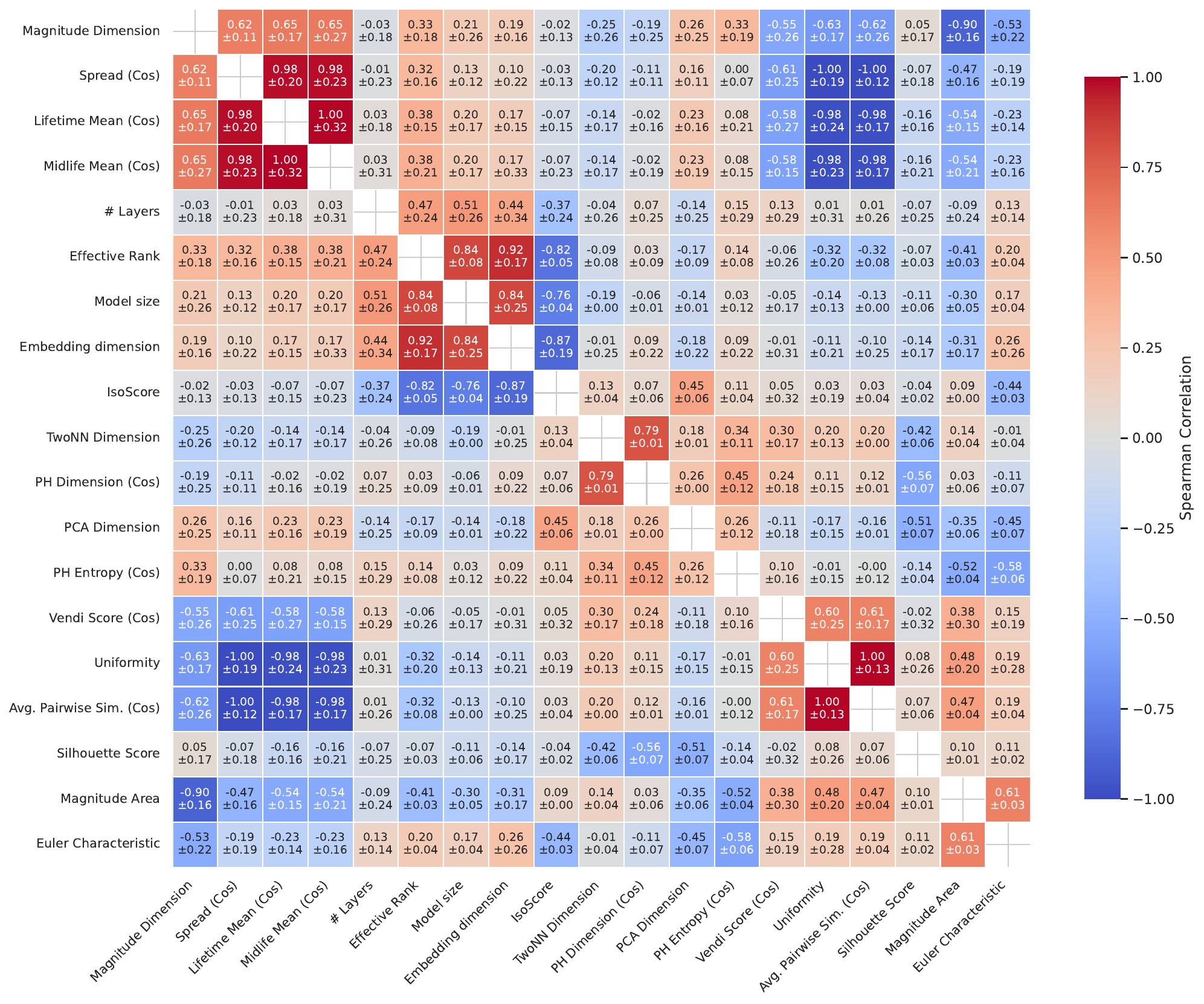}}
    \caption{Average correlations of topological descriptors and model properties.}
    \label{correlation}
    \end{center}
    \vskip -0.2in
\end{figure*}

We perform a correlation analysis and aggregate the pairwise correlations within each dataset among all used descriptors and additional properties such as the number of model layers and the embedding dimension. Figure \ref*{correlation} shows the mean correlation and variance between datasets, organized using hierarchical linkage clustering. We observe that IsoScore is highly (negatively) correlated with embedding dimension and effective rank, indicating that larger embedding dimensions lead to more anisotropic spaces. Given that persistent homology descriptors emerge from pairwise distances, we find that they are highly correlated with the pairwise similarity descriptor. Finally, some of the intrinsic dimension estimators are correlated with each other, indicating that they capture similar aspects of the data. These correlations allow to select a subset of descriptors that are less expensive to compute, while still capturing the overall topology.

\section{Representational similarity analysis}
\label{a_similairty}

\subsection{Centered Kernel Alignment}

\begin{figure}[h] 
    \vskip 0.2in
    \begin{center}
        \begin{minipage}{0.48\textwidth}
            \centerline{\includegraphics[width=1\columnwidth]{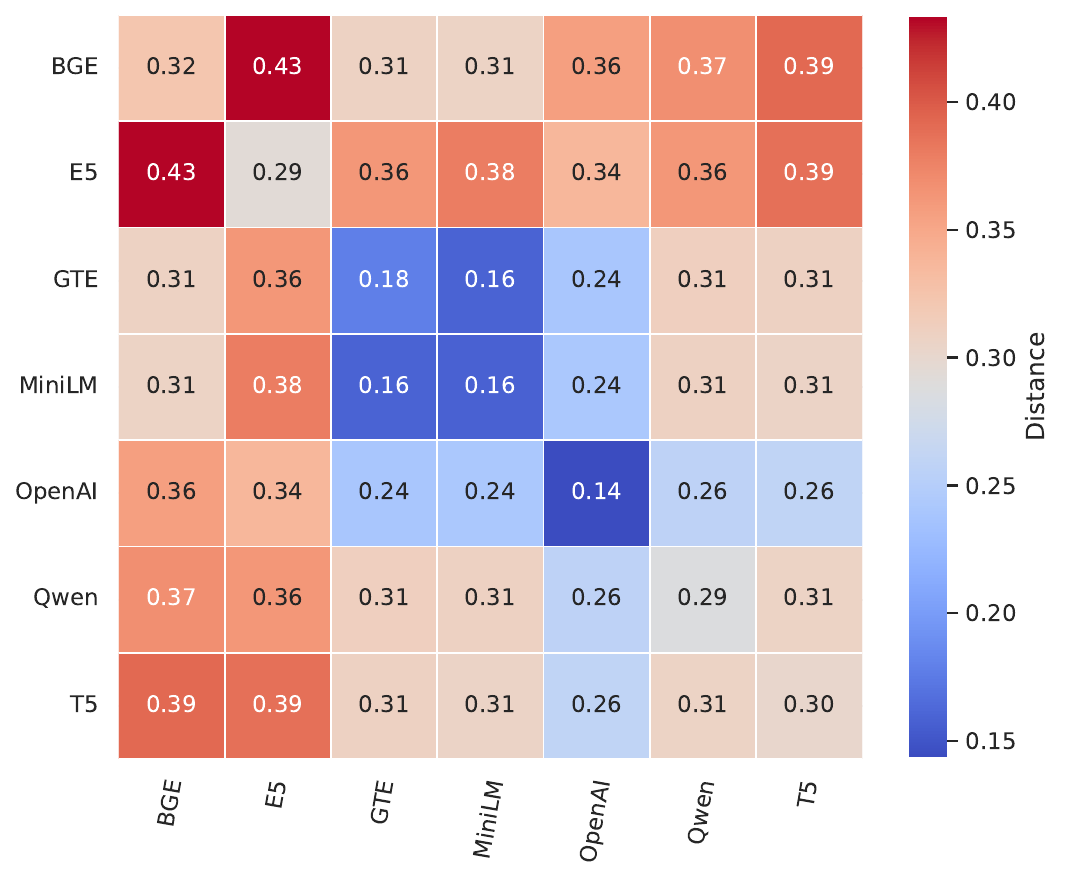}}
            \caption{Centered Kernel Alignment between model families.}
            \label{cka}
        \end{minipage}%
        \hfill
        \begin{minipage}{0.48\textwidth}
            \centerline{\includegraphics[width=1\columnwidth]{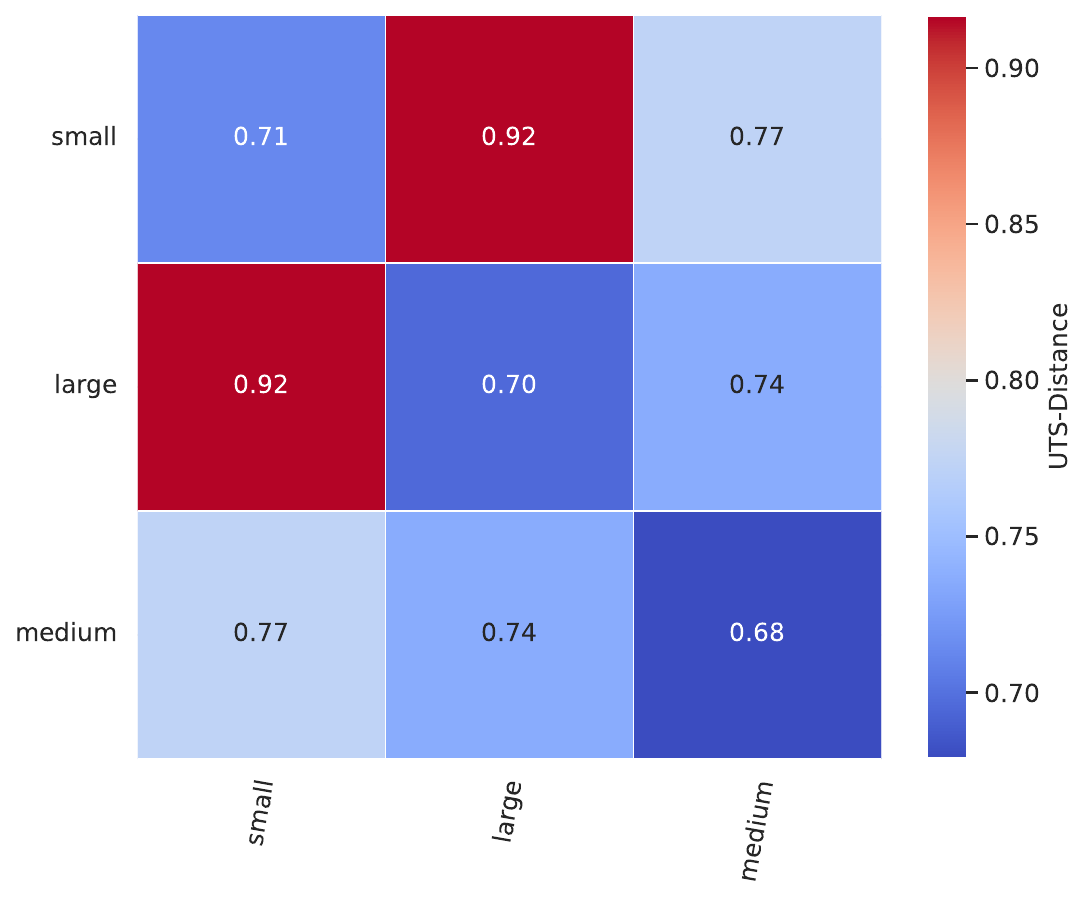}}
            \caption{UTS-based similarity of different model sizes.}
            \label{sizes}
        \end{minipage}
    \end{center} 
    \vskip -0.2in 
\end{figure}

We compute the Centered Kernel Alignment between all embedding models and datasets and aggregate the average alignment as shown in Figure \ref*{cka}. We find a similar pattern as with our UTS-based similarity measure, emphasizing that the model architecture and training data play a significant role in the alignment of embedding model representations. 

\subsection{Model size}
By computing UTS similarity, we find that models of the same size are more similar to each other than to models of different sizes. Furthermore, the topologies of large and small models are most different, which can be explained by the dissimilar transformer architectures.

\begin{figure}[h]
    \vskip 0.2in
    \begin{center}
    \centerline{\includegraphics[width=0.8\columnwidth]{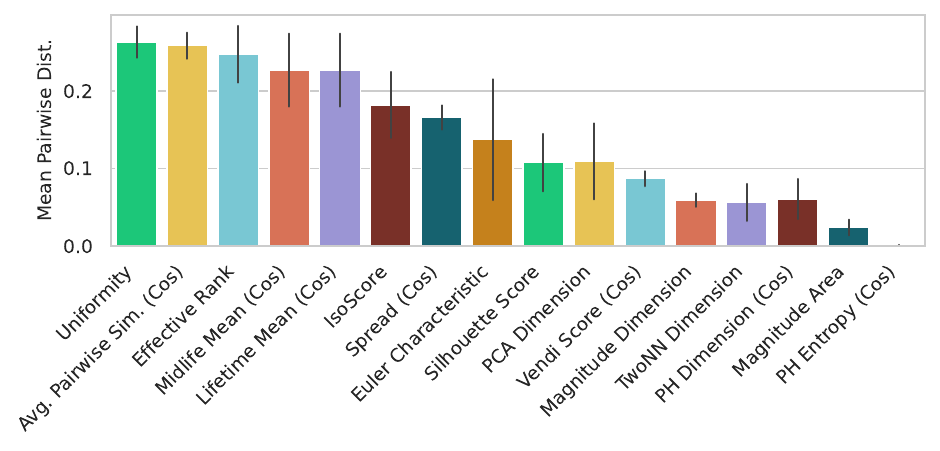}}
    \caption{Average distance of UTS components across all models.}
    \label{discriminative}
    \end{center}
    \vskip -0.2in
\end{figure} 

\subsection{Interpretability of similarity}
Our UTS approach enables the interpretability of representational similarity by inspecting the topological drivers that make representations converge. On one hand, we can determine the most discriminative features across all models, as summarized by Figure \ref*{discriminative}. We find that representations mostly converge due to their density (measured by uniformity and pairwise similarities) as well as their dimensionality (measured by IsoScore and effective rank). Furthermore, similarity is partially driven by the clustering of the embedding spaces. Alongside global interpretations, our framework highlights similarities across embedding spaces by analyzing pairwise comparisons, for example, showing that one space is denser while another is more isotropic. In the context of the Platonic Representation Hypothesis, we specifically find that each model family has different drivers for representational alignment. For instance, Qwen models have a significantly higher effective rank than all other models, while the Clustering coefficient is most discriminative for Gemini models.

\section{Influence of dimensionality on retrieval performance}
\label{a_retrieval}

\begin{figure}[t]
    \vskip 0.2in
    \begin{center}
    \centerline{\includegraphics[width=\columnwidth]{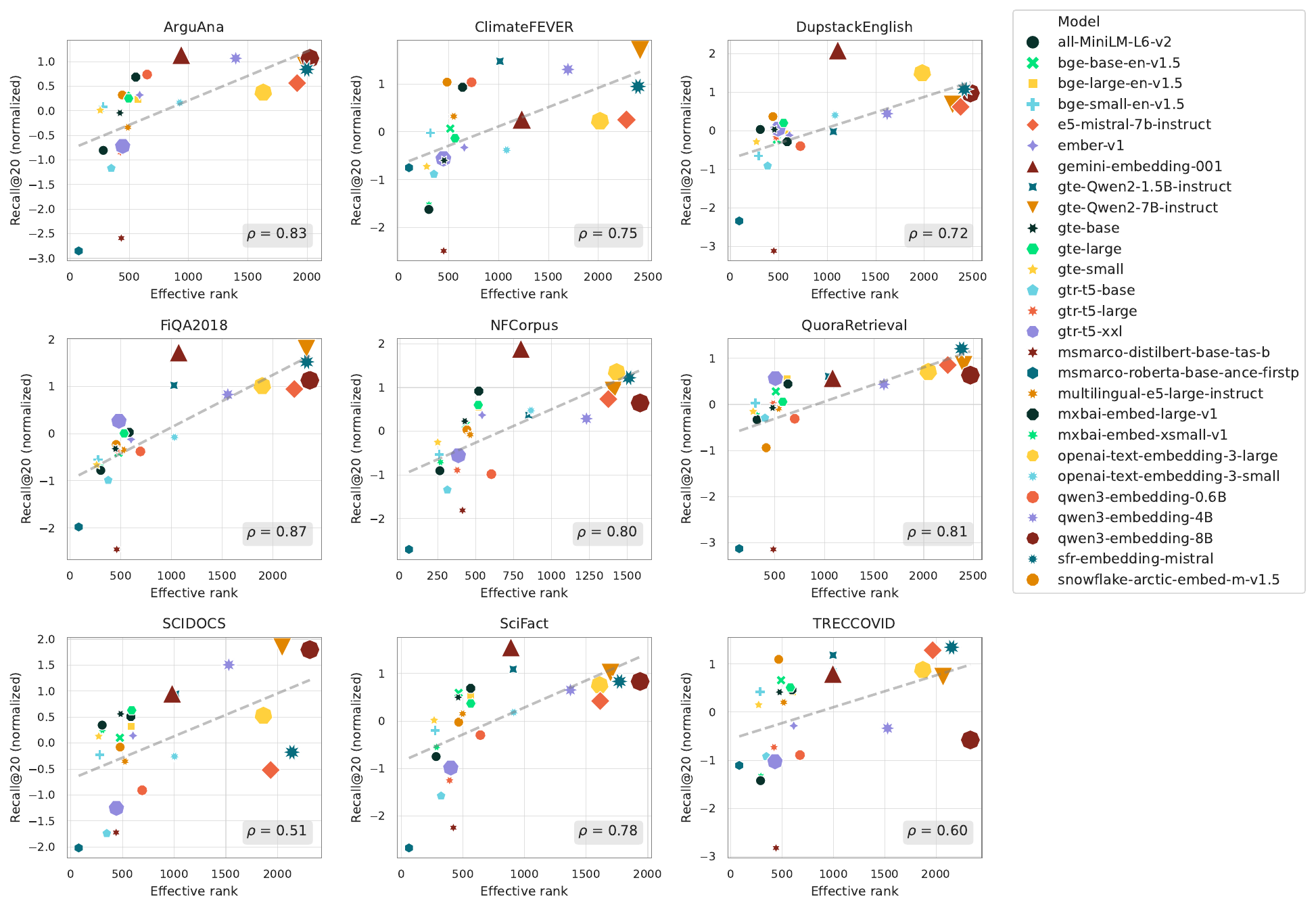}}
        \caption{Relationship between normalized Recall@20 and effective rank for different datasets. }
    \label{retrieval_er}
    \end{center}
    \vskip -0.2in
\end{figure}

As shown in Figure~\ref*{retrieval_er}, the effective rank exhibits consistently strong correlations 
with retrieval performance metrics, such as Recall@20. To visualize this relationship, we encode the 
embedding model size by marker size and observe that larger models generally correspond to higher 
effective ranks. This observation aligns with the well known trend that model scale correlates with performance. However, in our setting, a plausible confounding factor is the embedding 
dimension, which has been shown to play a critical role in retrieval systems~\cite{weller2025theoretical}. 
Indeed, our correlation analysis demonstrates that effective rank is itself highly correlated with the 
embedding dimension, suggesting that the observed relationship between model size and effective rank 
is largely attributed to this relationship.

\section{Predictive analysis}
\label{a_predictive}

\begin{table*}[t]
    \caption{\textbf{Balanced accuracy} for predicting model-specific
    properties from topological signatures across all folds, controlling
    for embedding dimensionality. \#Cls: number of classes; BL: majority
    class baseline.}
    \label{pred-cv}
    \small
    \renewcommand\arraystretch{1.3}
    \resizebox{\textwidth}{!}{%
    \begin{tabular}{l | ccc | ccc | ccc | ccc}
    \toprule
    & \multicolumn{3}{c|}{Model}
    & \multicolumn{3}{c|}{Model family}
    & \multicolumn{3}{c|}{Architecture}
    & \multicolumn{3}{c}{Activation} \\
    \cmidrule(lr){2-4}\cmidrule(lr){5-7}\cmidrule(lr){8-10}\cmidrule(lr){11-13}
    Dim. & Acc. & \#Cls & BL
         & Acc. & \#Cls & BL
         & Acc. & \#Cls & BL
         & Acc. & \#Cls & BL \\
    \midrule
    All
        & $0.66 \pm 0.02$ & 27 & 0.04
        & $0.68 \pm 0.07$ & 12 & 0.17
        & $0.98 \pm 0.01$ &  2 & 0.72
        & $0.84 \pm 0.12$ &  3 & 0.59 \\[2pt]
    \hline
    \\[-8pt]
    4096
        & $0.89 \pm 0.10$ &  3 & 0.37
        & $1.00 \pm 0.00$ &  2 & 0.73
        & $1.00 \pm 0.00$ &  1 & 1.00
        & $1.00 \pm 0.00$ &  1 & 1.00 \\
    1024
        & $0.68 \pm 0.23$ &  5 & 0.20
        & $0.68 \pm 0.23$ &  5 & 0.20
        & $0.78 \pm 0.26$ &  2 & 0.80
        & $0.78 \pm 0.26$ &  2 & 0.80 \\
    768
        & $0.73 \pm 0.12$ &  8 & 0.13
        & $0.83 \pm 0.15$ &  6 & 0.38
        & $1.00 \pm 0.00$ &  1 & 1.00
        & $0.86 \pm 0.06$ &  2 & 0.63 \\
    384
        & $0.76 \pm 0.01$ &  4 & 0.25
        & $0.93 \pm 0.13$ &  2 & 0.50
        & $1.00 \pm 0.00$ &  1 & 1.00
        & $1.00 \pm 0.00$ &  1 & 1.00 \\
    \bottomrule
    \end{tabular}}
\end{table*}


\subsection{Architectural attributes}
We train classifiers to  predict the model architecture (encoder vs. decoder) and the activation function (GELU, ReLU, SiLU) used in unseen embedding spaces. We find that both properties can be predicted with a higher balanced accuracy than the baseline, as summarized in Table \ref*{pred-cv}. For architecture, the binary classification appears to be an easy task, driven by the large difference in embedding dimension between encoder and decoder models. Therefore, we again control for embedding dimension but find that there are too few models of different architectures with the same embedding dimension. For the activation functions, we find a similar pattern suggesting that the same architecture mostly uses the same activation function. Despite this, we are able to predict the architecture used in an embedding spaces with a balanced accuracy of $0.98 \pm 0.01$, compared to a baseline of $0.72$, and the activation function with a balanced accuracy of $0.84 \pm 0.12$, compared to a baseline of $0.59$. To simulate predictions unseen models, we perform a group cross-validation, yielding comparable results ($0.94 \pm 0.11$ for architecture and $0.83 \pm 0.12$ for activation function). These results suggest that \gls*{uts} capture model-specific details encoded in the global topology of the embedding space, motivating further research into understanding how specific architectural choices affect the geometry of embedding spaces, similar to recent work on LayerNorm \cite{gupta2024geometric}.

\subsection{PCA signatures}
To reduce redundancy and uncover the most informative dimensions of variation, we apply \gls*{pca} \cite{pearson1901pca} to the collection of normalized signature vectors across all embedding spaces. 
This yields a reduced signature $\hat{\boldsymbol{s}}^{X} = [\hat{s}_1^{X}, \hat{s}_2^{X}, \dots, \hat{s}_L^{X}]^\top  \quad (L < K)$, where each component \( \hat{s}_i^{X} \) is a linear combination of the original normalized descriptors, capturing the principal axes of variation in the topological feature space. We analyze the effectiveness of the PCA-transformed signature vectors using only the first 5 principal components as features. We find that the average accuracies are slightly lower with 0.44 for predicting the model, 0.50 for predicting model family. This suggests that highly correlated attributes might still complement each other in prediction tasks. In the case of retrieval performance prediction, we also observe lower correlations compared to the full UTS vectors, approaching the results of effective rank. This can be explained by the large influence of properties related to the embedding dimension, which are captured by the first principal component. 

\subsection{Multilingual datasets}
We conduct an additional experiment on the GerDaLIRSmall dataset (queries=12k, docs=10k), which contains domain-specific law documents in German language. We compute retrieval metrics and topological signatures for 3 multilingual models of the Qwen family (Qwen2-1.5B-instruct, Qwen3-0.6B, Qwen3-4B) across 3 seeds. We train a classifier to predict the multilingual model that generated an embedding space using the signatures of other datasets as training data. We then predict on GerDaLIRSmall as test set and find that we can correctly classify 6 out of 9 embedding spaces (i.e. balanced accuracy = 66.66 vs. 33.33 as baseline). Note that this prediction task is especially difficult, because the models origin from the same family, sharing architectural similarities, reflected in similar signatures. This additional experiment shows that our approach can be directly applied on non-English datasets, but might benefit from having more multilingual training data.

\section{Signature clustering analysis}
\begin{figure}[h]
    \begin{center}
    \centerline{\includegraphics[width=\columnwidth]{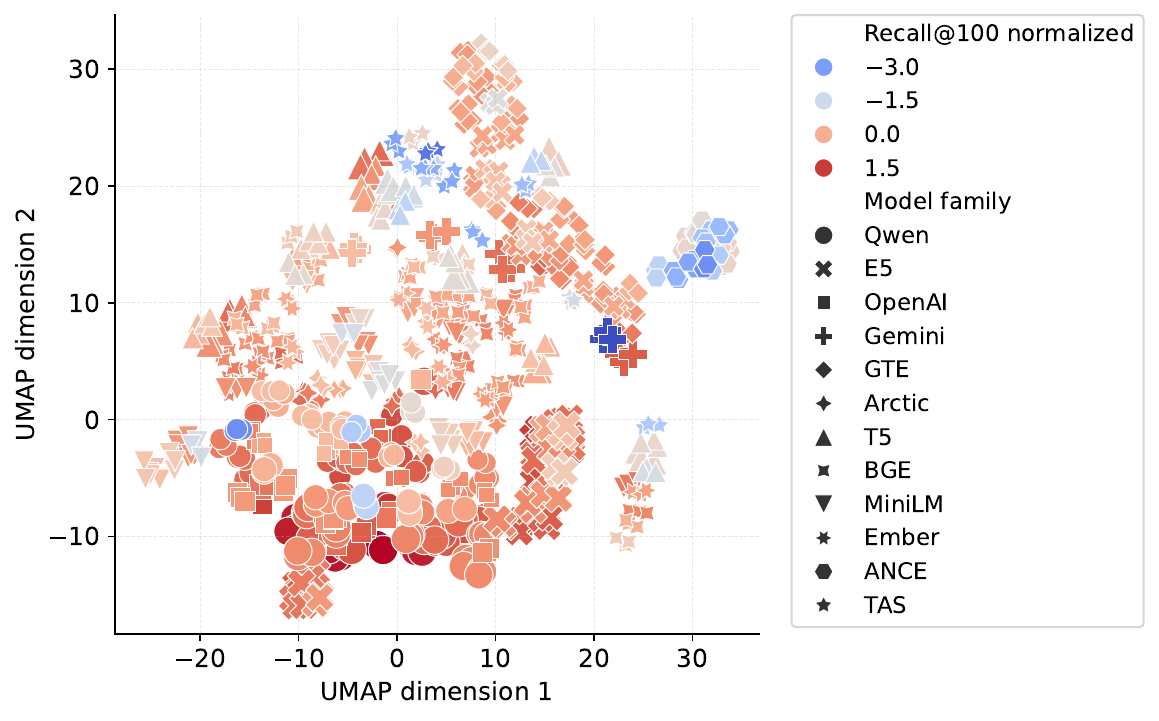}}
    \caption{Clustering of Unified Topological Signatures and their normalized Recall@100.}
    \label{sig_clust}
    \end{center}
\end{figure}

We apply UMAP to the signature vectors of all models, seeds and datasets 
shown in Figure \ref*{sig_clust}. We encode the model family by different markers, the retrieval performance (normalized Recall@100) by colors and the model size by the size of the points. First, we find that models cluster by model family, which is in line with our previous findings. Moreover, proprietary models developed by OpenAI cluster within the same region as decoder-based models (such as Qwen and E5), highlighting the potential to infer architectural characteristics from our holistic signatures. We also observe a general clustering by model size, which correlates with architecture, given that larger models are typically decoder-based. Finally, we find that sub-clusters correspond to retrieval performance, suggesting that embedding spaces with comparable topology yield similar retrieval outcomes.

\section{Global retrievability analysis}
\label{a_retrievability_global}
\begin{figure}[h]
    \vskip 0.2in
    \begin{center}
    \centerline{\includegraphics[width=0.9\columnwidth]{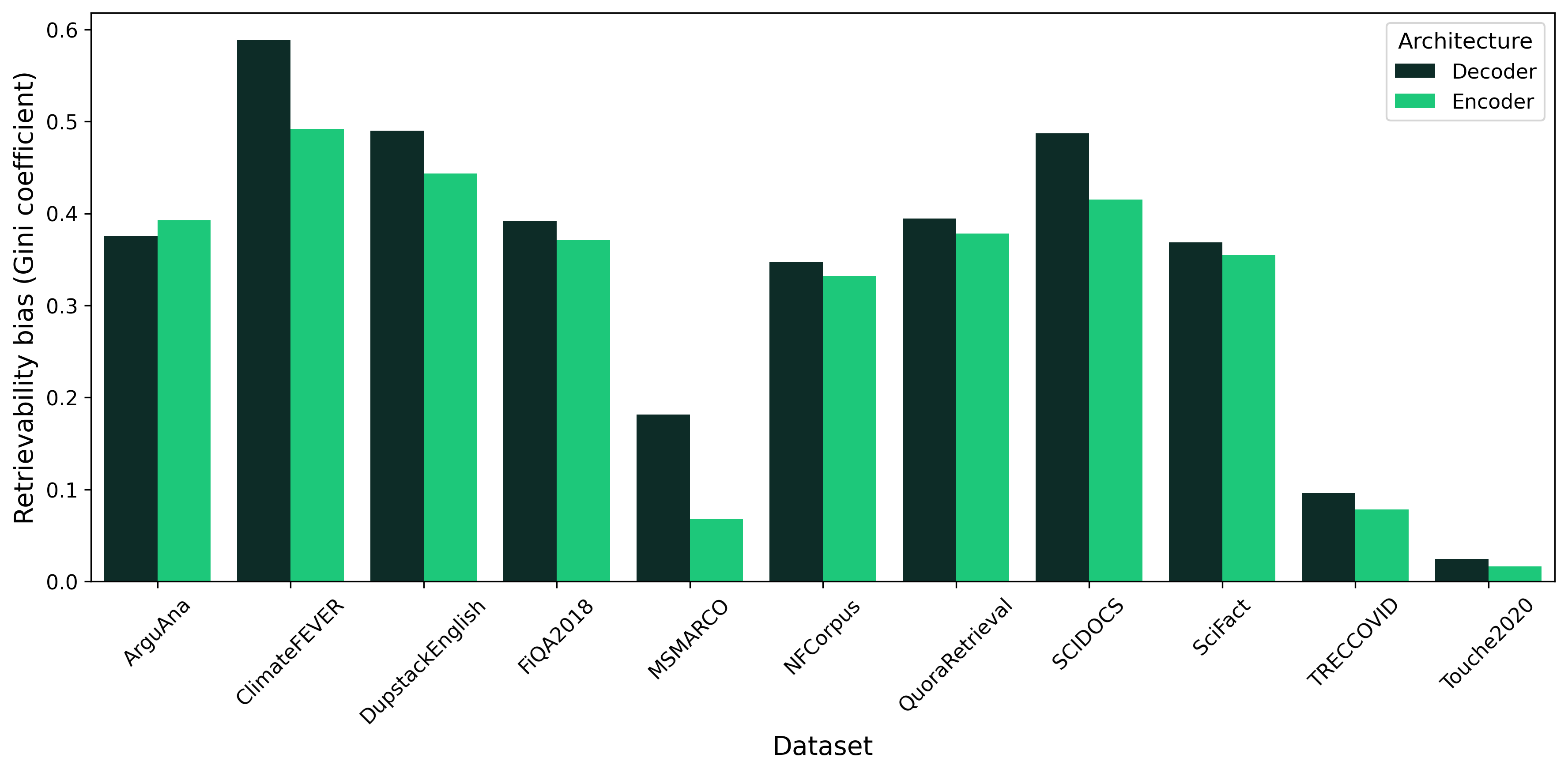}}
    \caption{Retrievability bias of encoder and decoder models on different datasets.}
    \label{gini_retr}
    \end{center} 
    \vskip -0.2in
\end{figure}

We study the Gini coefficient of the top-100 retrievability distribution to assess the overall bias in a dense retrieval system. As displayed in Figure \ref*{gini_retr}, the retrieval datasets differ in bias strength, which is related to the collection size. Given a fixed query distribution, we observe that decoder-based models, on average, exhibit a larger bias across most datasets. This implies that they return certain documents more frequently than others. A key difference between the decoder and encoder models used in our experiments is their embedding dimension. We hypothesize, that these effects can be explained by the dimensionality affecting the similarity computation. This is a possible connection to previous studies that have analyzed the influence of dimensionality on retrieval \cite{reimers2021curse}. These findings suggest that not only retrieval performance is associated with embedding dimensionality, but also retrieval bias.

\section{Local retrievability analysis}
\label{a_retrievability}

\subsection{Predictions with raw embeddings}
\begin{figure}[ht] 
    \vskip 0.2in
    \begin{center}
        \begin{minipage}{0.48\textwidth}
            \centerline{\includegraphics[width=\columnwidth]{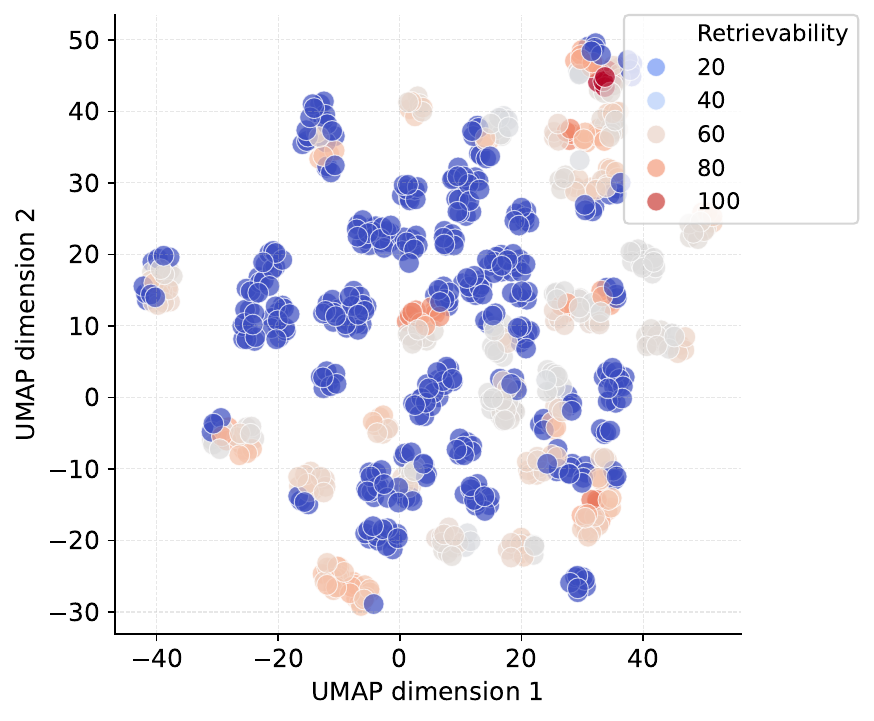}}
            \caption{ArguAna raw embeddings for text-embedding-3-large.}
            \label{raw1}
        \end{minipage}%
        \hfill
        \begin{minipage}{0.48\textwidth}
            \centerline{\includegraphics[width=\columnwidth]{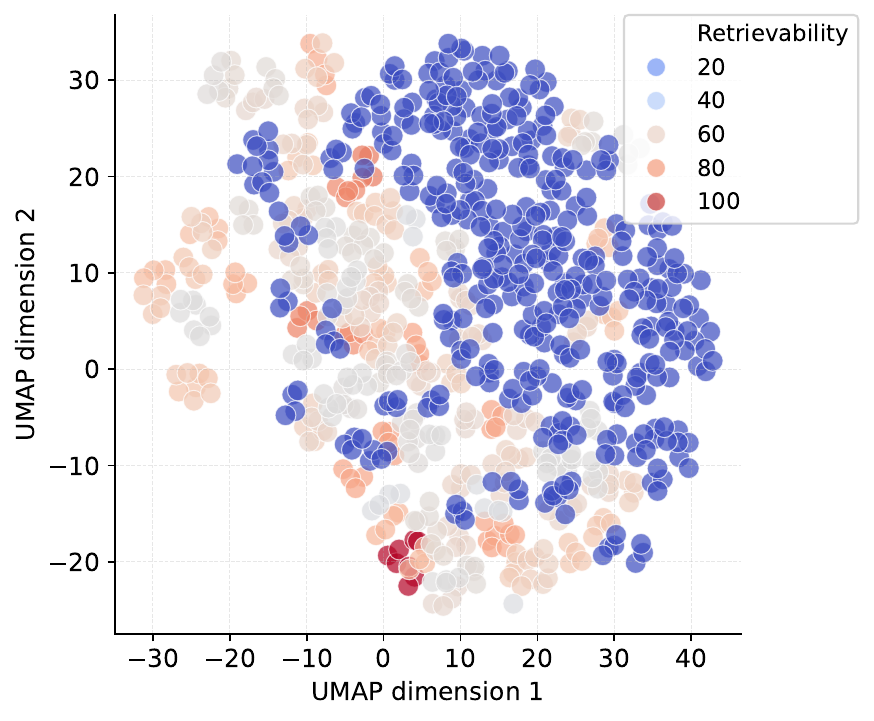}}
            \caption{ArguAna local UTS for text-embedding-3-large.}
            \label{uts1}
        \end{minipage}

        \vspace{0.3cm}
        
        \begin{minipage}{0.48\textwidth}
            \centerline{\includegraphics[width=\columnwidth]{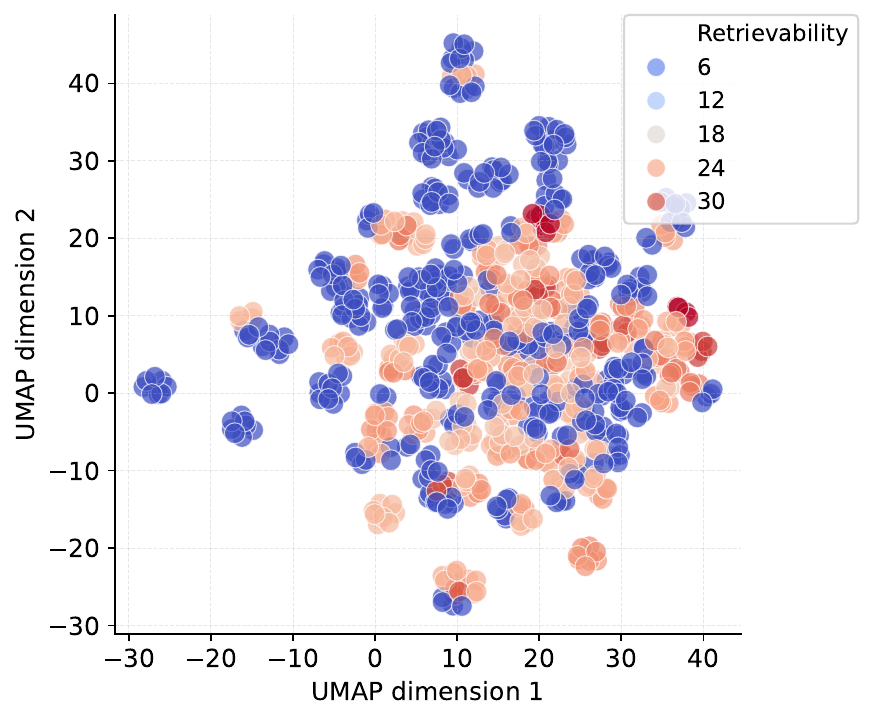}}
            \caption{NFCorpus raw embeddings for e5-mistral-7b-instruct.}
            \label{raw2}
        \end{minipage}%
        \hfill
        \begin{minipage}{0.48\textwidth}
            \centerline{\includegraphics[width=\columnwidth]{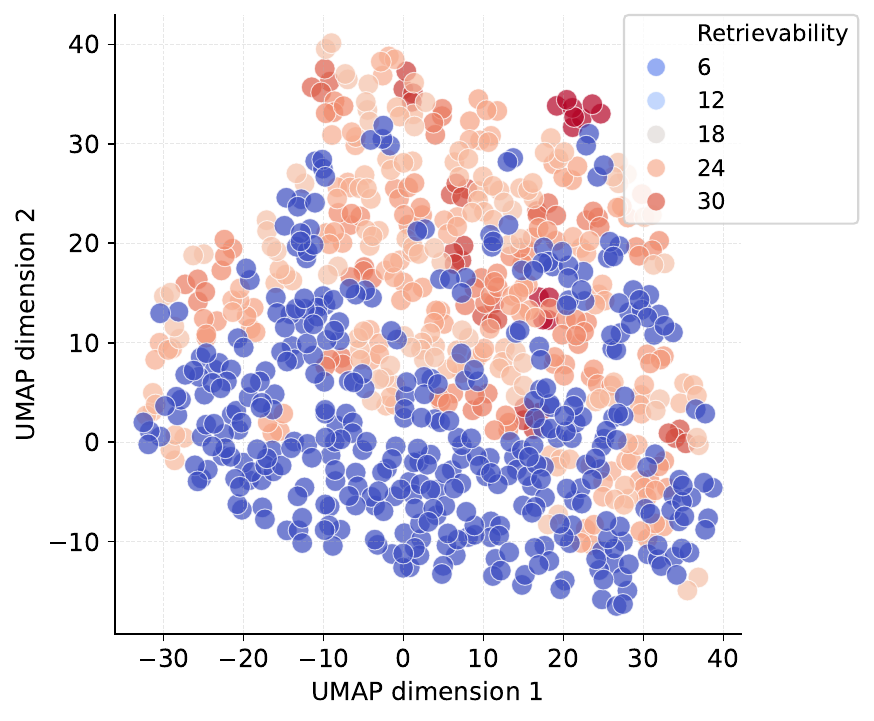}}
            \caption{NFCorpus local UTS for e5-mistral-7b-instruct.}
            \label{uts2}
        \end{minipage}

        \vspace{0.3cm}

        \begin{minipage}{0.48\textwidth}
            \centerline{\includegraphics[width=\columnwidth]{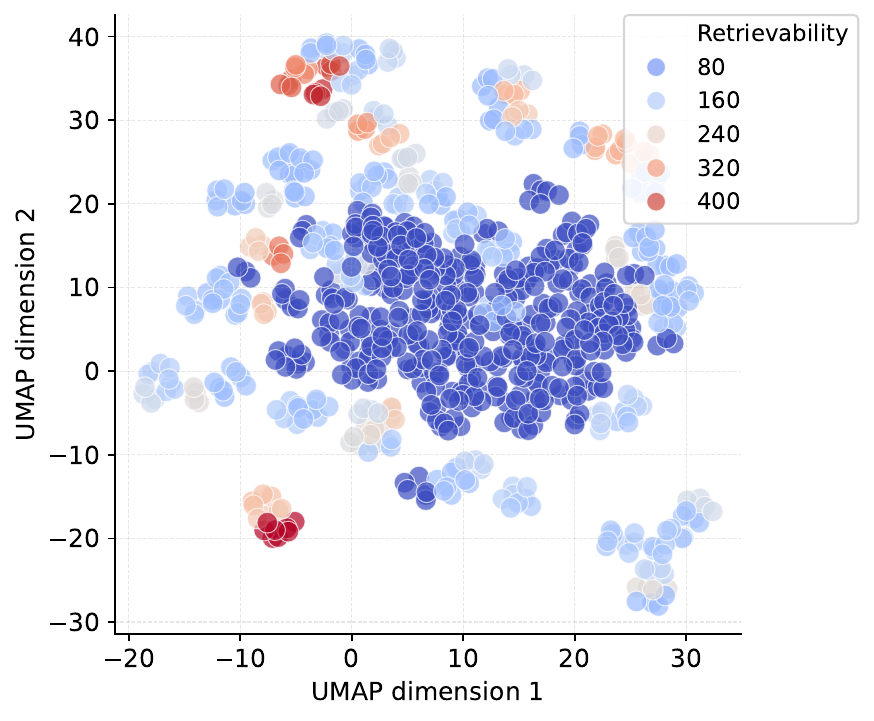}}
            \caption{Quora raw embeddings for multilingual-e5-large.}
            \label{raw3}
        \end{minipage}%
        \hfill
        \begin{minipage}{0.48\textwidth}
            \centerline{\includegraphics[width=\columnwidth]{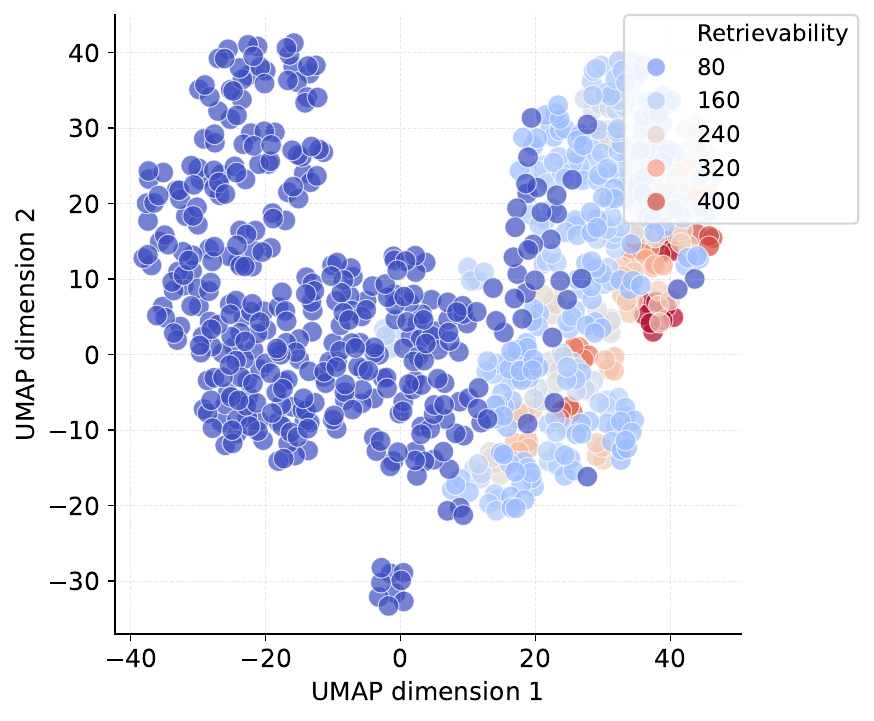}}
            \caption{Quora local UTS for multilingual-e5-large.}
            \label{uts3}
        \end{minipage}
    \label{comparisons}
    \end{center} 
    \vskip -0.2in 
\end{figure}

As our results have shown, the topological signatures reveal significant differences between retrievable and non-retrievable documents. We investigate whether such separations exist in the raw embedding space, as this would eliminate the need of computing signatures. For this, we use UMAP to visualize the raw embedding space and compare it with the UTS space, color-coding the retrievability of documents. As Figures \ref*{raw1}--\ref*{uts3} show, the UTS space provides a much clearer separation than the raw embedding space, motivating our approach. These findings are supported by the lower accuracy of the baseline prediction model, as reported in the results section. Finally, even if the raw embedding space shows a partial separation (e.g. Figure \ref*{raw3}), such trends do not extrapolate well to embedding spaces of other datasets, unlike our proposed UTS vectors.

\subsection{Characteristics of retrievable documents}
\begin{figure}[ht] 
    \vskip 0.2in
    \begin{center}
    \centerline{\includegraphics[width=\columnwidth]{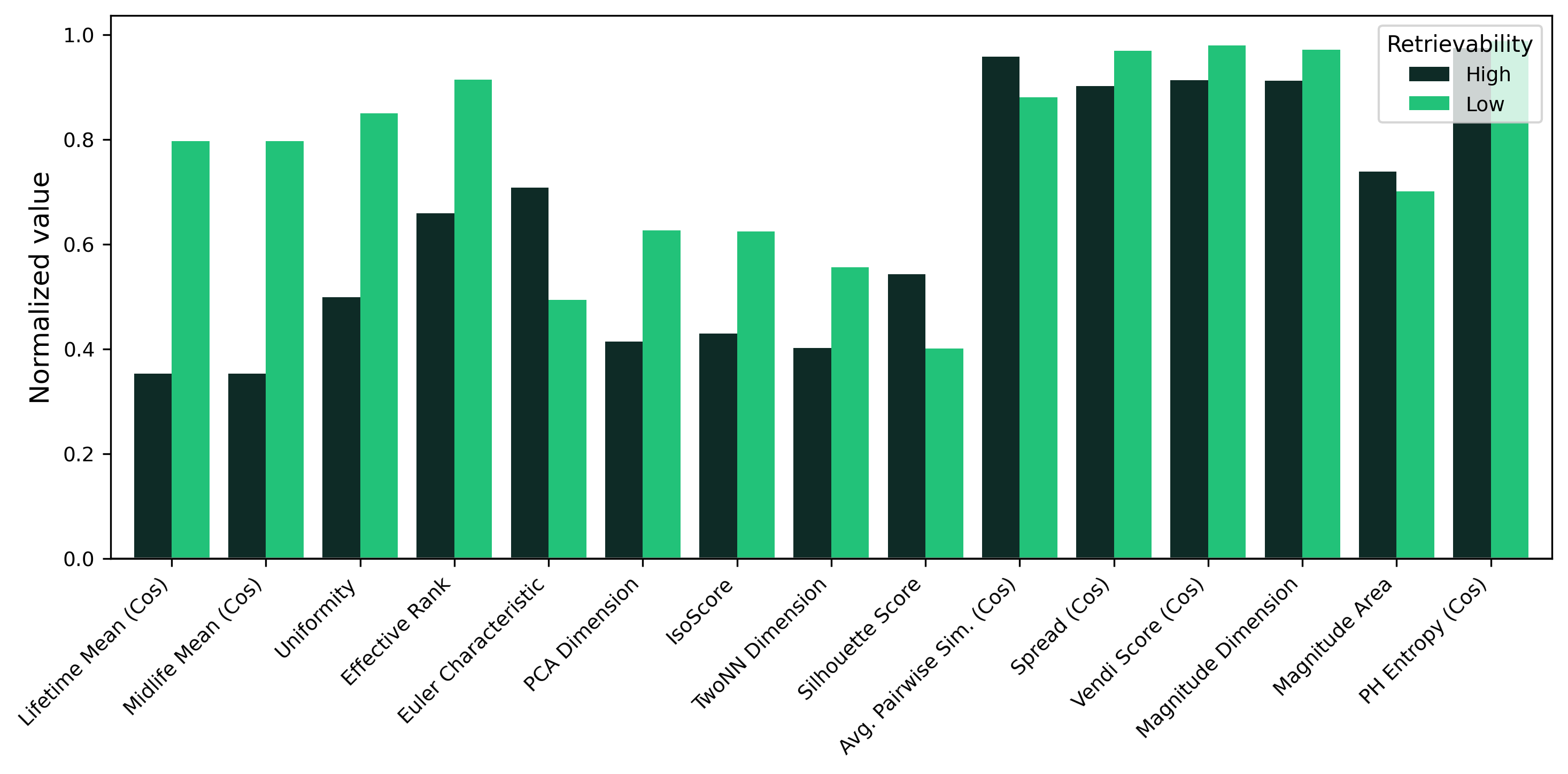}}
    \caption{Differences between clusters of highly retrievable and non-retrievable documents for QuoraRetrieval dataset.}
    \label{differences}
    \end{center} 
    \vskip -0.2in 
\end{figure}

\begin{figure}[ht] 
    \vskip 0.2in
    \begin{center}
    \centerline{\includegraphics[width=\columnwidth]{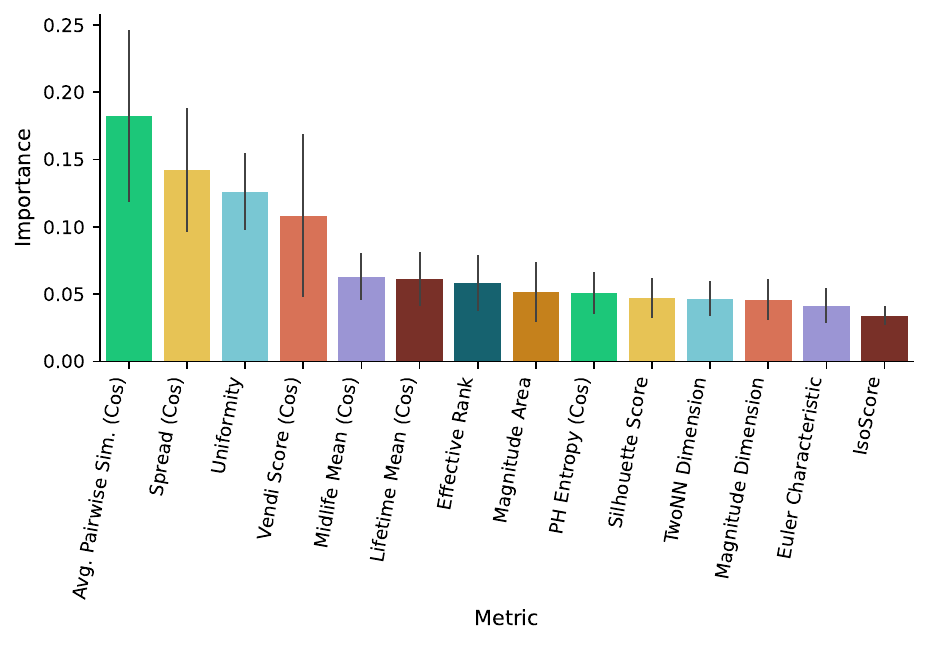}}
    \caption{Feature importance for predicting retrievability.}
    \label{retrievability_importance}
    \end{center} 
    \vskip -0.2in 
\end{figure}

To gain deeper insight into the topological features that influence document retrievability, we analyze the differences between retrievable and non-retrievable clusters. For this, we select 10 data points from each of the clusters and compare their topological signatures. As Figure \ref*{differences} shows, the local topology has distinct values for several descriptors, highlighting the importance of our unified approach. Further, we find that the largest differences lie in properties related to the density of the embedding space, suggesting that less retrievable documents lie in isolated regions. Beyond density, we also find that the neighborhood of highly retrievable documents has lower intrinsic dimensions. These observations are generalized by the feature importance analysis of the prediction model as summarized by Figure \ref*{retrievability_importance}.

\subsection{Retrievability of embedding models}
\begin{figure}[t]
    \centerline{\includegraphics[width=\columnwidth]{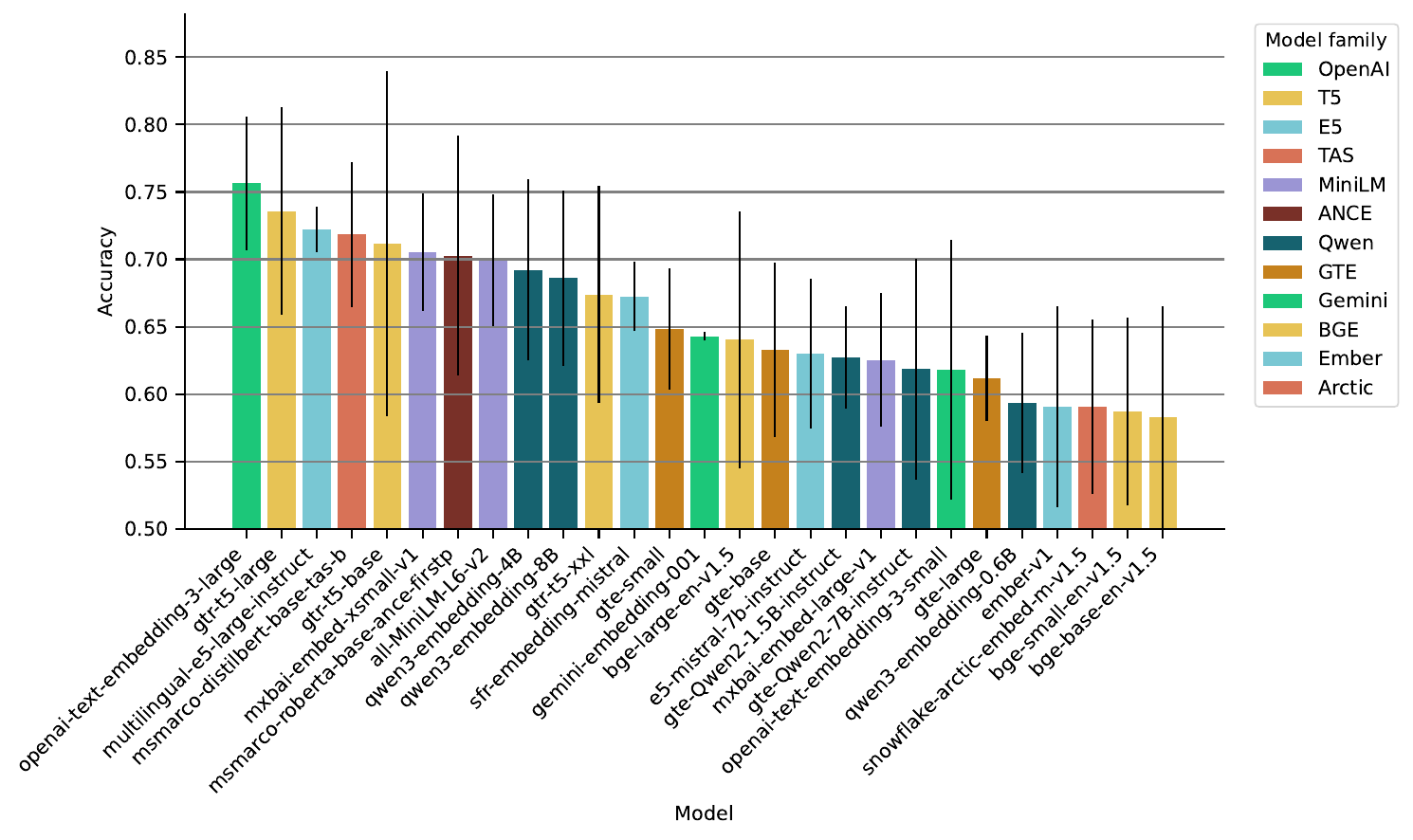}}
    \caption{Comparison of accuracies by different models for predicting retrievability.}
    \label{model_retr}
\end{figure}

We investigate whether embeddings differ in their ability to predict document retrievability. For this, we fit prediction models for each embedding model and evaluate the accuracy using group cross-validation. As Figure \ref*{model_retr} shows, there are large differences in the accuracy between different models, with OpenAI's text-embedding-3-large achieving the best results. This suggests that the structural characteristics of local topologies vary across models, motivating further exploration of retrievability.

\section{Ablation studies}
\label{a:ablations}
We perform a series of ablation studies to evaluate the robustness of our method. 

\paragraph{Descriptor sensitivity} First, we evaluate the effect of removing descriptors from the signature, to evaluate how the prediction results change. As shown in Table \ref*{tab:descriptor_ablation}, the full signature consistently achieves the best overall performance, while density and isotropy descriptors contribute most strongly to predictive performance. The last row corresponds to the baseline performance when using only effective rank as feature.

\paragraph{Normalization strategy} Furthermore, in Table \ref*{tab:normalization_ablation}, we investigate the robustness when using different normalization strategies and find that all methods maintain similar performance. To complement these results, we also provide the minimum, mean, and maximum values in Table \ref*{tab:descriptor_ranges} for each descriptor used in the UTS signature. The observed ranges indicate that the descriptors do not exhibit extreme outliers, supporting the robustness of the normalization procedure.

\paragraph{Model choice} Lastly, we evaluate different prediction models in Table~\ref*{tab:model_comparison} to assess the influence of classifier choice on downstream performance. Overall, model choice has a measurable but limited impact on the results. Across all prediction tasks, the Random Forest consistently achieves the highest balanced accuracy and is therefore used throughout our experiments. While the MLP performs competitively on some tasks, it generally underperforms the Random Forest and exhibits no clear advantage. The Nearest Neighbor classifier yields the weakest overall results, suggesting that the relationship between topological descriptors and model properties cannot be captured solely through local similarity.

\begin{table}[h] \caption{Balanced accuracy (mean $\pm$ standard deviation across cross-validation folds) for predicting model-specific properties after successively removing descriptor groups. } \label{tab:descriptor_ablation} \small \renewcommand\arraystretch{1.3} \centering \begin{tabular}{l | c | c | c | c} \toprule & \textbf{Model} & \textbf{Model Family} & \textbf{Architecture} & \textbf{Activation} \\ \midrule \textbf{All} & $\mathbf{0.66 \pm 0.02}$ & $\mathbf{0.68 \pm 0.07}$ & $\mathbf{0.98 \pm 0.01}$ & $\mathbf{0.84 \pm 0.10}$ \\ \midrule Remove PH descriptors & $0.64 \pm 0.02$ & $0.67 \pm 0.07$ & $0.98 \pm 0.01$ & $0.82 \pm 0.07$ \\ Remove ID descriptors & $0.62 \pm 0.02$ & $0.67 \pm 0.07$ & $0.98 \pm 0.01$ & $0.84 \pm 0.09$ \\ Remove Density descriptors & $0.41 \pm 0.01$ & $0.40 \pm 0.07$ & $0.98 \pm 0.01$ & $0.70 \pm 0.20$ \\ Remove Isotropy descriptors & $0.39 \pm 0.01$ & $0.47 \pm 0.07$ & $0.98 \pm 0.01$ & $0.78 \pm 0.20$ \\ \bottomrule \end{tabular} \end{table}

\begin{table}[h] \caption{Balanced accuracy (mean $\pm$ standard deviation across cross-validation folds) obtained using different normalization procedures.} \label{tab:normalization_ablation} \small \renewcommand\arraystretch{1.3} \centering \begin{tabular}{l | c | c | c | c} \toprule & \textbf{Model} & \textbf{Model Family} & \textbf{Architecture} & \textbf{Activation} \\ \midrule \textbf{Max normalization} & $\mathbf{0.66 \pm 0.02}$ & $\mathbf{0.68 \pm 0.07}$ & $\mathbf{0.98 \pm 0.01}$ & $\mathbf{0.84 \pm 0.10}$ \\ \midrule Mean normalization & $0.66 \pm 0.04$ & $0.68 \pm 0.09$ & $0.98 \pm 0.01$ & $0.83 \pm 0.05$ \\ Robust scaler & $0.65 \pm 0.04$ & $0.68 \pm 0.09$ & $0.98 \pm 0.01$ & $0.84 \pm 0.05$ \\ Standard scaler & $0.65 \pm 0.04$ & $0.68 \pm 0.09$ & $0.98 \pm 0.01$ & $0.84 \pm 0.06$ \\ \bottomrule \end{tabular} \end{table}

\begin{table}[h] \caption{Impact of classifier choice on downstream prediction tasks. Balanced accuracy (mean $\pm$ standard deviation across cross-validation folds) obtained using different prediction models.} \label{tab:model_comparison} \small \renewcommand\arraystretch{1.3} \centering \begin{tabular}{l | c | c | c | c} \toprule & \textbf{Model} & \textbf{Model Family} & \textbf{Architecture} & \textbf{Activation} \\ \midrule \textbf{Random Forest} & $\mathbf{0.66 \pm 0.02}$ & $\mathbf{0.68 \pm 0.05}$ & $\mathbf{0.98 \pm 0.01}$ & $\mathbf{0.84 \pm 0.10}$ \\ \midrule MLP & $0.57 \pm 0.04$ & $0.66 \pm 0.05$ & $0.96 \pm 0.01$ & $0.77 \pm 0.09$ \\ Nearest Neighbor & $0.49 \pm 0.07$ & $0.63 \pm 0.07$ & $0.93 \pm 0.04$ & $0.78 \pm 0.10$ \\ \bottomrule \end{tabular} \end{table}

\begin{table*}[t] \caption{Distribution of descriptor values across all embedding spaces.} \label{tab:descriptor_ranges} \small \renewcommand\arraystretch{1.2} \centering \begin{tabular}{l | r | r | r} \toprule \textbf{Descriptor} & \textbf{Min} & \textbf{Mean} & \textbf{Max} \\ \midrule Magnitude Area & 43.8468 & 1314.3146 & 1735.8951 \\ Magnitude Dimension & 2.2974 & 4.9628 & 5.6931 \\ PCA Dimension & 34.0000 & 118.6526 & 310.0000 \\ TwoNN Dimension & 4.7281 & 20.0096 & 37.0267 \\ Silhouette Score & 0.0059 & 0.0376 & 0.2335 \\ Isolation Score & 0.0000 & 0.1255 & 0.4551 \\ Uniformity & -3.9189 & -2.0565 & -0.0423 \\ Euler Characteristic & -20990.0 & -5368.3743 & 2324.0 \\ Effective Rank & 59.9828 & 814.8092 & 2572.4165 \\ Average Pairwise Cosine Similarity & 0.0077 & 0.4758 & 0.9894 \\ Mean Persistence Lifetime (Cosine) & 0.0037 & 0.2636 & 0.6032 \\ Mean Persistence Midlife (Cosine) & 0.0019 & 0.1318 & 0.3016 \\ Vendi Score (Cosine) & 1.0078 & 1.3272 & 1.4449 \\ Persistent Homology Dimension (Cosine) & 0.0000 & 10.7721 & 21.7708 \\ Persistent Homology Entropy (Cosine) & 6.4838 & 13.4932 & 14.2780 \\ Spread (Cosine) & 1.0106 & 1.7332 & 2.6893 \\ \bottomrule \end{tabular} \end{table*}

\clearpage

\end{document}